\pdfoutput=1
\documentclass[11pt]{article}
\usepackage[final]{acl}

\usepackage{times}
\usepackage{latexsym}
\usepackage{algorithm}
\usepackage{algorithmic}
\usepackage{amsmath}
\usepackage{booktabs}
\usepackage{multirow} 
\usepackage[table]{xcolor}
\usepackage{colortbl}
\definecolor{myskyblue}{RGB}{210, 235, 255}
\usepackage{textgreek}
\usepackage{tcolorbox} 
\usepackage[T1]{fontenc}
\usepackage[utf8]{inputenc}
\usepackage{inconsolata}
\usepackage{graphicx}
\usepackage{tikz}
\usetikzlibrary{positioning, arrows.meta, calc, shapes.geometric, fit, backgrounds}
 \usepackage{amssymb}
\usepackage{setspace}
\usepackage[patch=none]{microtype}

\title{SAGE: A Search-AuGmented Evaluation of Large Language Models \\
       on Free-Form QA}

\author{
  \textbf{Sher Badshah\textsuperscript{1}},
  \textbf{Ali Emami\textsuperscript{2}},
  \textbf{Hassan Sajjad\textsuperscript{1}} \\
  \textsuperscript{1}Dalhousie University \quad
  \textsuperscript{2}Emory University \\
  \texttt{\{sh545346, hsajjad\}@dal.ca} \quad
  \texttt{aemami@emory.edu}
}

\begin{document}
\maketitle
\begin{abstract}
As Large Language Models (LLMs) become increasingly used for question-answering (QA), relying on static, pre-annotated references for evaluation poses significant challenges in cost, scalability, and completeness. Meanwhile, using LLMs themselves as evaluators without external grounding remains unreliable for objective tasks, as they systematically over-accept incorrect answers, fabricate supporting rationales, and degrade sharply on questions that fall outside their training data. We propose Search-AuGmented Evaluation (SAGE), a framework to assess LLM outputs without fixed ground-truth answers. Unlike conventional metrics that compare to static references or depend solely on LLM-as-a-judge knowledge, SAGE acts as an agent that actively retrieves and synthesizes external evidence. It iteratively generates web queries, collects information, summarizes findings, and refines subsequent searches through reflection. By reducing dependence on static reference-driven evaluation protocols, SAGE offers a scalable and adaptive alternative for evaluating the factuality of LLMs. Experimental results on multiple free-form QA benchmarks show that SAGE achieves substantial to perfect agreement with human evaluations.
\end{abstract}

\section{Introduction}
\label{sec:intro}
Free-form Question Answering (QA) requires models to generate precise natural language responses to broad, open-ended queries~\citep{wang2023evaluating}. As such, it serves as a key benchmark for evaluating the factuality of Large Language Models (LLMs), which are increasingly integrated into real-world applications such as online search engines and virtual assistants. However, LLMs are prone to hallucination~\citep{gou2024criticllms}, and evaluating their factuality with standard protocols remains difficult.

Traditional evaluation methods, including lexical matching metrics such as Exact Match (EM) and F1, rely on comparisons to static ground-truth references. While convenient and efficient, these methods fall short of capturing the diversity of free-form QA outputs and are often infeasible to scale due to the high cost of human annotations~\citep{chianglee2023large, manas2024improving, zhu2023judgelm}. More critically, instruction-tuned LLMs produce outputs that are often unpredictable, context-dependent, and non-deterministic, making it impractical to pre-annotate reference answers for every possible response~\citep{yehudai2025surveyevaluationllmbasedagents, li2024dynaeval}. As a result, static, reference-driven evaluation protocols are fundamentally misaligned with the nature of free-form QA, where answers are open-ended and often lack a single definitive ground truth.

An emerging alternative reference-based evaluation is the LLM-as-a-judge approach~\citep{10.5555/3666122.3668142, chen2024mllm}, where one model, for instance, is prompted to assess the output of another based on task-specific criteria such as relevance, depth, or creativity~\citep{verga2024replacingjudgesjuriesevaluating}. This method has shown promise in subjective tasks such as summarization, dialogue, and instruction following, where quality is shaped by style or user preference and multiple interpretations are often equally valid~\citep{gu2025surveyllmasajudge, son2024llmasajudgerewardmodel}. However, its reliability deteriorates when the goal shifts to objective correctness~\citep{krumdick2025freelabelslimitationsllmasajudge, DAFE, gu2025surveyllmasajudge}. 

\begin{figure*}[t]
  \includegraphics[width=\textwidth]{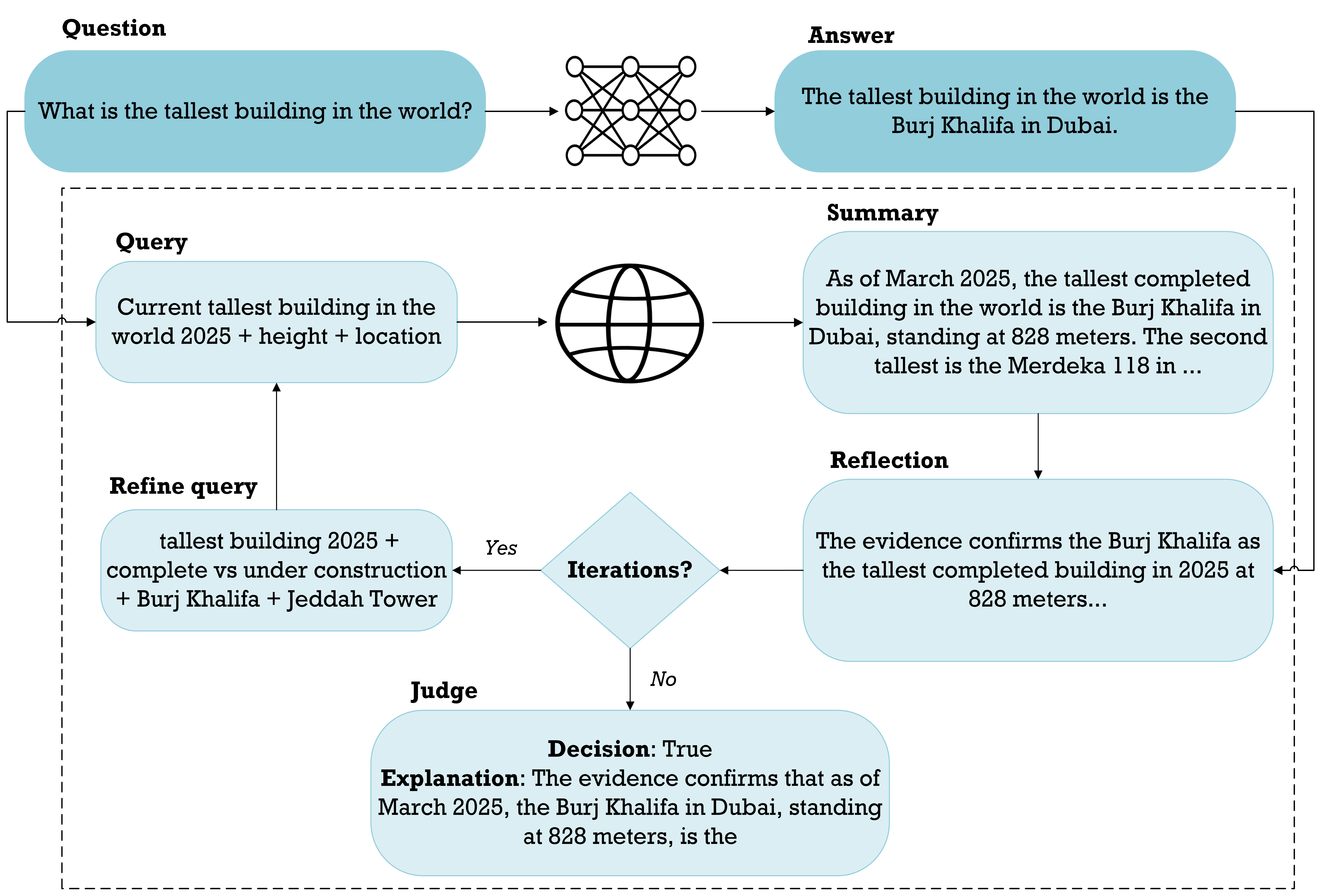}
  \caption{Given the question \textit{``What is the tallest building in the world?''} and candidate answer \textit{``The tallest building in the world is the Burj Khalifa in Dubai,''}, SAGE begins with \textbf{\textit{initial query}} from the question. The query triggers \textbf{\textit{web searches}} across multiple sources, followed by \textbf{\textit{evidence summarization}} to extract key insights. The \textbf{\textit{reflection module}} assesses evidence sufficiency and relevance, triggering \textbf{\textit{query refinement}} if needed. After N iterations, the \textbf{\textit{judge}} synthesizes the evidence to provide a final decision with rationale.} 
  \vspace{-4pt}
  \label{fig:sage_overview}
\end{figure*}

For objective, fact‑centric tasks such as free-form QA, an unguided (i.e., reference-free) judge is forced to lean solely on its frozen pre-trained knowledge. This constraint exposes several recurring failure modes: i) knowledge staleness where the judges confidently endorse answers that became outdated after their training cut off~\citep{vu-etal-2024-freshllms, cheng2024dateddatatracingknowledge, DAFE}; ii) length or verbosity bias, where longer or more detailed answers are overrated even when they contain errors~\citep{li2025evaluatingscoringbiasllmasajudge, ye2024justiceprejudicequantifyingbiases}; iii) prompt sensitivity, in which small variations to the prompt or the order of candidate answers flip a correct/incorrect verdict~\citep{ye2024justiceprejudicequantifyingbiases, thakur2025judgingjudgesevaluatingalignment}; and iv) hallucinated rationales, where the judge develops supporting evidence to justify its decision~\citep{kamalloo-etal-2023-evaluating}.

Given that pre‑annotated reference answers at scale are impractical and reference‑free LLM‑as‑a‑judge setups remain largely unreliable for objective tasks, we argue that, unlike subjective evaluation, objective correctness cannot be assessed solely through an LLM’s parametric knowledge or its preferences. Therefore, we propose Search-AuGmented Evaluation (SAGE), a novel framework that bridges the stated gap by equipping LLM judges with the ability to actively collect and synthesize external evidence. Instead of costly human-annotated reference answers, SAGE dynamically and iteratively generates output-specific web queries, retrieves information, reflects on findings, and refines its search strategy to verify the correctness of model outputs. 

SAGE offers four key advantages: (1) it substantially reduces dependency on the judge's parameter knowledge, (2) avoids the need for human-annotated reference answers, making evaluation more scalable, (3) grounds evaluations in up to date, verifiable information, and (4) enables assessment of novel or rapidly evolving topics where parameter knowledge may be outdated or incomplete. Through experiments on free-form QA, we find that SAGE is aligned with reference-based evaluation and achieves substantial to perfect agreement with human evaluators.
\section{Methodology}
\label{sec:method}
We introduce Search-AuGmented Evaluation (SAGE), a reference-free framework for evaluating LLM responses. Unlike conventional approaches that rely on fixed reference answers or human-annotated ground truths, SAGE autonomously gathers and integrates external evidence to assess the correctness of free-form responses. Figure~\ref{fig:sage_overview} illustrates the overall process.

Let $x$ denote an input question and let $C$ be a candidate LLM that produces a response $\hat{y} = C(x)$. Our goal is to decide, without access to a pre-annotated reference answer, whether $\hat{y}$ is factually correct. SAGE addresses this by equipping an LLM judge $J$ to iteratively retrieve and reason over external evidence. Internally, $J$ plays four roles via distinct one-shot prompt templates: query generation ($Q$), summarization ($\Sigma$), reflection ($\Phi$), and final judgment. These roles are composed with an external web-search tool ($S$) and an append-only short-term memory buffer ($\mathcal{M}$) that records the trace of their interaction. We describe each component below.

\paragraph{Query Generation ($Q$):}
At iteration $i$, a query generator $Q$ conditions on the question $x$ and the short-term memory $\mathcal{M}_{i-1}$ accumulated in prior iterations:
\begin{equation}
q_i = Q(x, \mathcal{M}_{i-1}).
\end{equation}
When $\mathcal{M}_{i-1} = \varnothing$ (i.e., $i=1$), $Q$ produces an initial, topic-level query from $x$ alone, without assuming the correctness of $\hat{y}$. For $i>1$, $Q$ refines the query using the accumulated evidence and reflections in $\mathcal{M}_{i-1}$, targeting information still needed to verify $\hat{y}$.\footnote{The initial and refinement prompts are templated variants of the same module; see Figures~\ref{fig:query_generation_prompt} and~\ref{fig:query_refinement_prompt} in the Appendix.} SAGE runs for a fixed budget of $N$ iterations; adaptive, reflection-driven stopping is discussed in Section~\ref{sec:limitations}.

\paragraph{Web Search ($S$):} The query is submitted to a web search engine via the Serper API,\footnote{\url{https://serper.dev/}} which returns real-time results. $S(q_i)$ denotes the set of up to $k=3$ snippets (title, text, and source URL) returned for $q_i$; these raw results serve as the external evidence consumed by the next step.

\paragraph{Summarization ($\Sigma$):} The retrieved results are condensed into a focused evidence segment,
\begin{equation}
E_i = \Sigma\bigl(S(q_i)\bigr),
\end{equation}
where $\Sigma$ extracts salient factual content and filters redundant or irrelevant text. The resulting $E_i$ is a concise, interpretable evidence summary used by the reflection step.

\paragraph{Reflection ($\Phi$):} The summarized evidence is assessed for relevance, sufficiency, and factual alignment with respect to $x$ and $\hat{y}$. This step yields a reflection
\begin{equation}
r_i = \Phi(x, \hat{y}, E_i),
\end{equation}
which records whether the current evidence supports, contradicts, or is inconclusive with respect to $\hat{y}$, and flags missing information or ambiguities that should guide the next query. The reflection output is what drives query refinement in the following iteration.

\paragraph{Short-Term Memory ($\mathcal{M}$):} We maintain an ordered buffer $\mathcal{M}$ whose state at the end of iteration $i$ is the sequence
\begin{equation}
\mathcal{M}_i = \bigl((q_1, E_1, r_1), \ldots, (q_i, E_i, r_i)\bigr),
\end{equation}
with $\mathcal{M}_0 = \varnothing$. The update is append-only: $\mathcal{M}_i = \mathcal{M}_{i-1} \,\|\, (q_i, E_i, r_i)$, where $\|$ denotes sequence concatenation. $\mathcal{M}$ records the trace of a single $(x, \hat{y})$ episode: it conditions query refinement in $Q$ and is passed in full to the judge at the end.

\paragraph{Judge ($J$):} After $N$ iterations, the judge is invoked on the full memory $\mathcal{M}_N$ and produces a binary verdict $v \in \{0,1\}$ together with a natural-language rationale $\rho$:
\begin{equation}
(v, \rho) = J(x, \hat{y}, \mathcal{M}_N).
\end{equation}
SAGE targets evidence-based verification of a single response: the judge asks whether the retrieved evidence supports or contradicts $\hat{y}$, not which of two responses is preferred nor how good $\hat{y}$ is on a scale. This is entailment-like by construction and distinct from pairwise preference judging or scalar quality rating; we therefore output a binary verdict and retain $\rho$ as the finer-grained reasoning trace for error analysis. The overall procedure is summarized in Algorithm~\ref{alg:SAGE}.

\begin{algorithm}[t]
\setstretch{1.3}
\caption{SAGE: Search-AuGmented Evaluation}
\label{alg:SAGE}
\begin{algorithmic}[1]
\REQUIRE question $x$, candidate response $\hat{y}$, iteration budget $N$
\ENSURE verdict $v\in\{0,1\}$, rationale $\rho$
\STATE $\mathcal{M} \leftarrow \varnothing$ \hfill\COMMENT{short-term memory}
\FOR{$i = 1$ \TO $N$}
    \STATE $q_i \leftarrow Q(x, \mathcal{M})$ \hfill\COMMENT{initial if $\mathcal{M}=\varnothing$, else refined}
    \STATE $E_i \leftarrow \Sigma\!\left(S(q_i)\right)$ \hfill\COMMENT{retrieve \& summarize}
    \STATE $r_i \leftarrow \Phi(x, \hat{y}, E_i)$ \hfill\COMMENT{reflect on sufficiency}
    \STATE $\mathcal{M} \leftarrow \mathcal{M} \,\|\, (q_i, E_i, r_i)$ \hfill\COMMENT{append to memory}
\ENDFOR
\STATE $(v, \rho) \leftarrow J(x, \hat{y}, \mathcal{M})$ \hfill\COMMENT{final verdict}
\RETURN $(v, \rho)$
\end{algorithmic}
\end{algorithm}
\section{Experimental Setup}
\label{sec:experiments}
\subsection{Models}
We use Gemini-1.5-pro~\citep{geminiteam2024gemini15unlockingmultimodal}, GPT-3.5-turbo~\citep{NEURIPS2020_1457c0d6}, and GPT-4o-mini~\citep{openai2023gpt4} as both candidates and judges within the SAGE framework, so that each model is assessed both on its ability to generate responses and on its ability to evaluate responses from others. All experiments use temperature $0$ to maximize determinism, as higher temperatures degrade LLM-based evaluators~\citep{hada2024llmbasedevaluators}. For brevity, we refer to these models as Gemini, GPT-3.5, and GPT-4o.

We further evaluate SAGE under smaller open-weight judges, including Mistral 7B~\citep{jiang2023mistral} and models from the Qwen family~\citep{qwen2025qwen25technicalreport, yang2025qwen3technicalreport} such as Qwen2.5-7B, Qwen3-4B, Qwen3-8B (see Appendix~\ref{app:small_llm}). 

\subsection{Datasets}
We evaluate SAGE on widely used free-form question-answering datasets that span different question types, knowledge domains, and complexity levels. These include AmbigQA~\citep{minetal2020ambigqa}, HotpotQA~\citep{yang2018hotpotqa}, TriviaQA~\citep{joshi2017triviaqa}, and Natural Questions (NQ-Open)~\citep{kwiatkowskietal2019}. Free-form question-answering underpins a broad range of practical applications in which accuracy and truthfulness are paramount~\citep{gou2024criticllms}. We also use FreshQA~\citep{vu2023freshllms} to evaluate SAGE's ability to detect outdated knowledge. Due to computational constraints, we randomly sample 300 instances per dataset. Each dataset provides reference answers that serve as ground truth for reference-based evaluators (see Appendix~\ref{app:datasets}).

\subsection{Prompts}
Our prompting strategy uses templates for both response generation and evaluation. For candidate models, we use few-shot Chain-of-Thought (CoT) prompts with 6 fixed exemplars per dataset to elicit detailed, reasoning-based responses, a strategy well-suited to free-form QA~\citep{gou2024criticllms}. In SAGE, we design module-specific prompts that combine role instructions with a step-by-step reasoning guide (see Appendix~\ref{app:prompting}).

\subsection{Baselines}
We compare SAGE against several established evaluation approaches, including reference-based and reference-free methods. Moreover, we conduct a human evaluation using two QA datasets. In the following, we summarize each baseline method. Appendix~\ref{app:baselines} provides further details on them.

\paragraph{Reference-based evaluation.} We consider \textit{\textbf{Exact Match (EM)}} and \textit{\textbf{F1}} as lexical reference-based baselines. Because standard automatic metrics can be misleading for free-form QA~\citep{badshah-etal-2025-clev, kamalloo-etal-2023-evaluating}, we also follow~\citet{badshah-sajjad-2025-reference} and~\citet{wang2023evaluating} and employ GPT-4 as a reference-based evaluator that compares candidate answers to gold answers; we refer to this evaluator as \textbf{RefGPT}. As shown in Table~\ref{tab:ref-based_agreement_human}, RefGPT attains consistently substantial agreement with human annotators, well above EM and F1. We therefore adopt RefGPT as our primary reference-based baseline. On datasets without human annotations, we further use RefGPT as a \emph{scalable proxy} (``silver human'') when computing SAGE's agreement, and we explicitly treat it as a proxy---not a gold standard: its $\kappa \approx 0.71$ with human judges indicates substantial but imperfect alignment, and our primary claims are anchored to the human-labeled subsets of AmbigQA and HotpotQA.

\begin{table}[t]
\centering
\scriptsize
\setlength{\tabcolsep}{3.5pt} 
\begin{tabular}{ll ccc ccc}
\toprule
\textbf{Candidate} & \textbf{Task} 
& \multicolumn{3}{c}{\textbf{Cohen's $\kappa$}} 
& \multicolumn{3}{c}{\textbf{Macro F1}} \\
\cmidrule(lr){3-5} \cmidrule(lr){6-8}
& & \textbf{EM} & \textbf{F1} & \textbf{RefGPT}
  & \textbf{EM} & \textbf{F1} & \textbf{RefGPT} \\
\midrule
\multirow{2}{*}{\textbf{GPT-3.5}} 
 & AmbigQA  & 0.54 & 0.66 & \textbf{0.76} & 0.76 & 0.83 & \textbf{0.88} \\
 & HotpotQA & 0.60 & 0.76 & \textbf{0.90} & 0.79 & 0.88 & \textbf{0.95} \\
\midrule
\multirow{2}{*}{\textbf{GPT-4o}}
 & AmbigQA  & 0.48 & 0.55 & \textbf{0.70} & 0.73 & 0.77 & \textbf{0.85} \\
 & HotpotQA & 0.54 & 0.66 & \textbf{0.77} & 0.76 & 0.83 & \textbf{0.88} \\
\midrule
\multirow{2}{*}{\textbf{Gemini}}
 & AmbigQA  & 0.56 & 0.57 & \textbf{0.71} & 0.77 & 0.78 & \textbf{0.85} \\
 & HotpotQA & 0.49 & 0.66 & \textbf{0.76} & 0.73 & 0.83 & \textbf{0.88} \\
\bottomrule
\end{tabular}
\caption{Agreement of reference-based metrics with human majority. F1 scores are converted to binary using a $\tau = 0.5$.}
\label{tab:ref-based_agreement_human}
\end{table}

\paragraph{Reference-free evaluation.} We adapt \textit{\textbf{Judge without search}} as a baseline, following the approach from~\citet{liu2023g}. In this setting, the judge relies entirely on its pre-trained knowledge to determine factual correctness. 

\paragraph{Human Evaluation.} We invite three graduate researchers to evaluate model outputs on AmbigQA and HotpotQA; due to budget constraints, human evaluation is limited to these two datasets. Annotators are presented with input questions, reference answers, and anonymized model responses in randomized order to prevent position or model-identity bias. Each response is rated on a binary scale: 1 (``True'') for responses that align with the reference answer and demonstrate contextual relevance, and 0 (``False'') otherwise. The majority vote determines the final judgment (see Appendix~\ref{app:human_eval}).

\subsection{Evaluation Metrics}
We report three metrics. \textit{\textbf{Accuracy}} is the proportion of instances where the judge's binary verdict matches the reference label obtained from automatic metrics or RefGPT. \textit{\textbf{Macro-F1}} measures a judge's agreement with reference-based metrics under class imbalance. For AmbigQA and HotpotQA, where human annotations are available, we additionally compute \textit{\textbf{Cohen's $\kappa$}} and \textit{\textbf{Macro-F1}} against the human majority vote. We further run ablations to quantify the \textit{\textbf{impact of specific SAGE components}}, measured as changes in agreement with human judgments.
\section{Results}
Our primary comparisons use RefGPT as a scalable proxy for human judgments on datasets without human annotations, and human majority votes on the AmbigQA and HotpotQA subsets where annotations are available. Additional results and ablations are reported in Appendices~\ref{app:additional_results} and~\ref{sec:add-ablations}.

\subsection{Main results}
\paragraph{External evidence improves agreement with reference-based evaluation.}
Table~\ref{tab:freej_sage_refgpt_acc} shows that SAGE agrees with RefGPT substantially more often than a parametric-only judge. For instance, GPT-3.5 as a SAGE judge reaches 0.80 accuracy when evaluating itself on AmbigQA, compared to only 0.67 without external evidence. The same pattern holds for GPT-4o and Gemini across all five datasets, indicating that grounding verdicts in retrieved evidence improves both precision and recall of the judge.

\paragraph{SAGE strongly agrees with human evaluations.}
To evaluate alignment with human judgment, we compare SAGE and baseline evaluators against majority votes from three expert annotators on AmbigQA and HotpotQA. As shown in Table~\ref{tab:k_mac_f1_freej_sage}, judges without search rely on their pre-trained knowledge, which often confirms the candidate's answer as correct. As a result, their agreement with human annotations is low, with Cohen's $\kappa$ often below 0.40. SAGE substantially closes this gap: for example, GPT-4o as a reference-free judge obtains $\kappa = 0.38$ on HotpotQA, whereas the same judge under SAGE reaches $\kappa = 0.70$. This pattern is notable because the judge model ``knows'' the correct answer as a candidate yet accepts contradictory claims at face value when acting as a reference-free judge---a failure mode that external evidence reliably corrects.

\begin{table}[t]
\centering
\scriptsize
\setlength{\tabcolsep}{1.6pt} 
\begin{tabular}{llccc ccc}
\toprule
\textbf{Cand.} & \textbf{Task} &
\multicolumn{3}{c}{\textbf{Judge without search}} &
\multicolumn{3}{c}{\textbf{SAGE}} \\
\cmidrule(lr){3-5}\cmidrule(lr){6-8}
& & \textbf{GPT‑3.5} & \textbf{GPT‑4o} & \textbf{Gemini} 
  & \textbf{GPT-3.5} & \textbf{GPT‑4o} & \textbf{Gemini} \\
\midrule
\multirow{5}{*}{\rotatebox[origin=c]{90}{\textbf{GPT‑3.5}}}
& AmbigQA  & 0.67 & 0.73 & 0.81 & 0.80 & \textbf{0.83} & 0.81 \\
& FreshQA  & 0.51 & 0.70 & 0.83 & 0.79 & \textbf{0.91} & 0.89 \\
& HotpotQA & 0.58 & 0.64 & 0.66 & 0.70 & \textbf{0.76} & 0.73 \\
& NQ‑Open  & 0.61 & 0.70 & 0.70 & 0.70 & 0.72 & \textbf{0.74} \\
& TriviaQA & 0.80 & 0.85 & 0.84 & 0.84 & \textbf{0.89} & 0.82 \\
\midrule
\multirow{5}{*}{\rotatebox[origin=c]{90}{\textbf{GPT‑4o}}}
& AmbigQA  & 0.70 & 0.70 & 0.79 & 0.80 & \textbf{0.83} & \textbf{0.83} \\
& FreshQA  & 0.54 & 0.59 & 0.68 & 0.76 & 0.78 & \textbf{0.81} \\
& HotpotQA & 0.57 & 0.63 & 0.62 & 0.70 & \textbf{0.77} & \textbf{0.77} \\
& NQ‑Open  & 0.59 & 0.65 & 0.71 & 0.71 & 0.74 & \textbf{0.75} \\
& TriviaQA & 0.82 & \textbf{0.86} & 0.81 & 0.84 & \textbf{0.86} & 0.80 \\
\midrule
\multirow{5}{*}{\rotatebox[origin=c]{90}{\textbf{Gemini}}}
& AmbigQA  & 0.68 & 0.72 & 0.70 & 0.75 & \textbf{0.83} & 0.76 \\
& FreshQA  & 0.64 & 0.64 & 0.65 & 0.73 & 0.75 & \textbf{0.76} \\
& HotpotQA & 0.61 & 0.63 & 0.61 & 0.75 & \textbf{0.76} & 0.75 \\
& NQ‑Open  & 0.61 & 0.62 & 0.61 & 0.64 & \textbf{0.72} & 0.67 \\
& TriviaQA & 0.81 & 0.82 & 0.79 & 0.83 & \textbf{0.85} & 0.80 \\
\bottomrule 
\end{tabular}
\caption{Agreement of Judge-without-search and SAGE with RefGPT across candidate models and tasks, measured as verdict accuracy against RefGPT's reference-based labels. Higher is better.}
\label{tab:freej_sage_refgpt_acc}
\end{table}

\begin{figure}[t]
    \centering
    \includegraphics[width=0.95\linewidth]{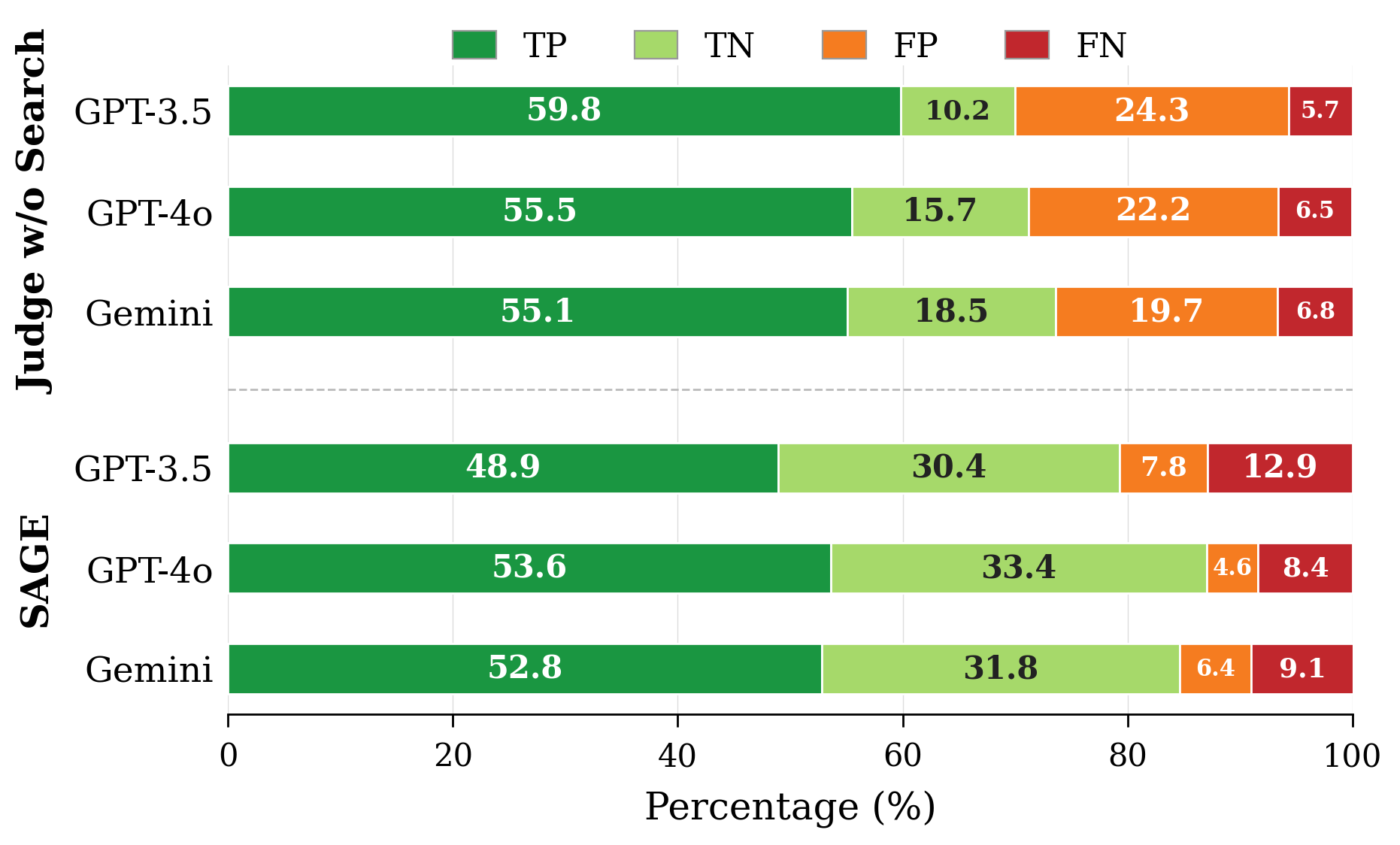}
    \caption{
        Distribution of true positive (TP), true negative (TN), false positive (FP), and false negative (FN) rates for each LLM judge in both Judge without search and SAGE, averaged across AmbigQA and HotpotQA and candidate models.}
    \label{fig:judge-method-confusion}
\end{figure}

Cohen’s $\kappa$ measures agreement beyond chance but can mislead under class imbalance, known as the \textit{kappa paradox}~\citep{CICCHETTI1990551}. Therefore, we report Macro F1, which treats both classes equally and provides a balanced view of evaluation performance. In Table~\ref{tab:k_mac_f1_freej_sage}, LLM-as-a-judge without access to reference answers shows competitive macro F1 scores, but analysis reveals a tendency to over-estimate correctness, leading to inflated recall at the expense of precision (see Figure~\ref{fig:judge-method-confusion}). In contrast, SAGE delivers the highest Macro F1 across models and tasks.

\begin{table*}[t]
\centering
\footnotesize
\setlength{\tabcolsep}{3.5pt}
\begin{tabular}{ll|cccccc|cccccc}
\toprule
\textbf{Candid.} & \textbf{Task} &
\multicolumn{6}{c}{\textbf{Judge without search}} &
\multicolumn{6}{c}{\textbf{SAGE}} \\
\cmidrule(lr){3-8} \cmidrule(lr){9-14}
& & \multicolumn{2}{c}{\textbf{GPT-3.5}} & \multicolumn{2}{c}{\textbf{GPT-4o}} & \multicolumn{2}{c}{\textbf{Gemini}} 
  & \multicolumn{2}{c}{\textbf{GPT-3.5}} & \multicolumn{2}{c}{\textbf{GPT-4o}} & \multicolumn{2}{c}{\textbf{Gemini}} \\
\cmidrule(lr){3-4} \cmidrule(lr){5-6} \cmidrule(lr){7-8}
\cmidrule(lr){9-10} \cmidrule(lr){11-12} \cmidrule(lr){13-14}
& & \textbf{$\kappa$} & \textbf{Mac-F1} & \textbf{$\kappa$} & \textbf{Mac-F1} & \textbf{$\kappa$} & \textbf{Mac-F1} 
  & \textbf{$\kappa$} & \textbf{Mac-F1} & \textbf{$\kappa$} & \textbf{Mac-F1} & \textbf{$\kappa$} & \textbf{Mac-F1} \\
\midrule
\multirow{2}{*}{\textbf{GPT-3.5}} 
 & AmbigQA  & 0.23 & 0.61 & 0.39 & 0.70 & 0.58 & 0.79 & 0.57 & 0.78 & \textbf{0.80} & \textbf{0.90} & 0.66 & 0.83 \\
 & HotpotQA & 0.16 & 0.53 & 0.26 & 0.61 & 0.35 & 0.67 & 0.41 & 0.71 & 0.56 & \textbf{0.78} & 0.53 & 0.76 \\
\midrule
\multirow{2}{*}{\textbf{GPT-4o}}
 & AmbigQA  & 0.24 & 0.59 & 0.38 & 0.68 & 0.57 & 0.78 & 0.60 & 0.80 & \textbf{0.91} & \textbf{0.96} & \textbf{0.91} & \textbf{0.96} \\
 & HotpotQA & 0.17 & 0.53 & 0.38 & 0.67 & 0.36 & 0.68 & 0.54 & 0.77 & 0.70 & \textbf{0.85} & 0.68 & 0.84 \\
\midrule
\multirow{2}{*}{\textbf{Gemini}}
 & AmbigQA  & 0.20 & 0.58 & 0.35 & 0.67 & 0.26 & 0.61 & 0.64 & 0.82 & 0.75 & \textbf{0.87} & 0.67 & 0.84 \\
 & HotpotQA & 0.17 & 0.55 & 0.27 & 0.62 & 0.27 & 0.62 & 0.63 & 0.82 & 0.69 & \textbf{0.85} & 0.59 & 0.80 \\
\bottomrule
\end{tabular}
\caption{Cohen's $\kappa$ and Macro-F1 between the human majority vote and reference-free judges (with and without SAGE) on AmbigQA and HotpotQA.}
\label{tab:k_mac_f1_freej_sage}
\end{table*}

\paragraph{SAGE works better with more capable judges.} SAGE's performance improves significantly when the judge is a more capable model, where ``more capable'' is defined by public leaderboard performance (e.g., MMLU, GSM8K) rather than by parameter count, which is undisclosed for several of the models we evaluate. In Table~\ref{tab:k_mac_f1_freej_sage}, GPT-4o outperforms GPT-3.5 and Gemini on both AmbigQA and HotpotQA; for example, GPT-4o as judge reaches a Macro-F1 of 0.90 on AmbigQA when evaluating GPT-3.5, higher than GPT-3.5's 0.78 and Gemini's 0.83.

\paragraph{SAGE detects untruthful facts and outdated knowledge.}
SAGE flags false claims by cross-referencing them with retrieved evidence. For \textit{``Who sings the theme song for the show Half \& Half?''}, a candidate model answered \textit{``Erica Campbell''} and tool-free judges accepted the claim; SAGE retrieved sources confirming that \textit{``Melonie Daniels''} performed the theme song and correctly rejected the candidate's answer with a grounded rationale. On FreshQA~\citep{vu2023freshllms}, SAGE similarly catches stale knowledge: for \textit{``Where is EMNLP this year?''}, candidates often returned outdated locations, while SAGE retrieved the correct current location \textit{``Suzhou, China''}\footnote{These experiments were conducted in 2025, when EMNLP was held in Suzhou, China; EMNLP 2026 is hosted in Hungary.} (see Appendix~\ref{app:untruthful_facts}).

\paragraph{SAGE fixes incorrect reasoning traces.} We analyzed cases where candidate models produced logically inconsistent or unsupported reasoning. 
SAGE’s reflection enables it to detect these inconsistencies (see Table~\ref{tab:incorrect_reasoning_example} in the Appendix).

\subsection{Error analysis}\label{sec:error_analysis}
To understand SAGE's limitations, we conducted a manual error analysis on 100 randomly sampled evaluation cases from AmbigQA and HotpotQA in which SAGE disagreed with the human majority. We group the errors into five categories: 1)~\textbf{Contextual misunderstanding (23\%):} SAGE generates inaccurate or incomplete queries when it misinterprets the candidate question's intent. This is particularly evident in AmbigQA, where questions are often intentionally ambiguous or under-specified, leading to irrelevant or contradictory evidence. 2)~\textbf{Incomplete evidence (15\%):} SAGE fails when the retrieved evidence is insufficient or lacks relevant information, especially for recent events with limited online coverage. 3)~\textbf{Reasoning error (32\%):} despite accurate evidence, the judge model misinterprets the information or applies flawed reasoning. 4)~\textbf{Hallucination (13\%):} when evidence is ambiguous or inconclusive, the judge falls back on its pre-trained knowledge and produces hallucinated rationales. 5)~\textbf{Conflicting evidence (7\%):} In some cases, SAGE encounters conflicting evidence across multiple search iterations. The framework is designed to iteratively refine its understanding; however, judges sometimes over-rely on earlier sources or fail to appropriately weigh the of conflicting information (details in Appendix~\ref{app:sage_failure}).

\subsection{Ablation study}
\begin{figure*}[t]
\begin{center}
\begin{minipage}{0.40\textwidth}
  \fbox{\includegraphics[width=\linewidth,keepaspectratio]{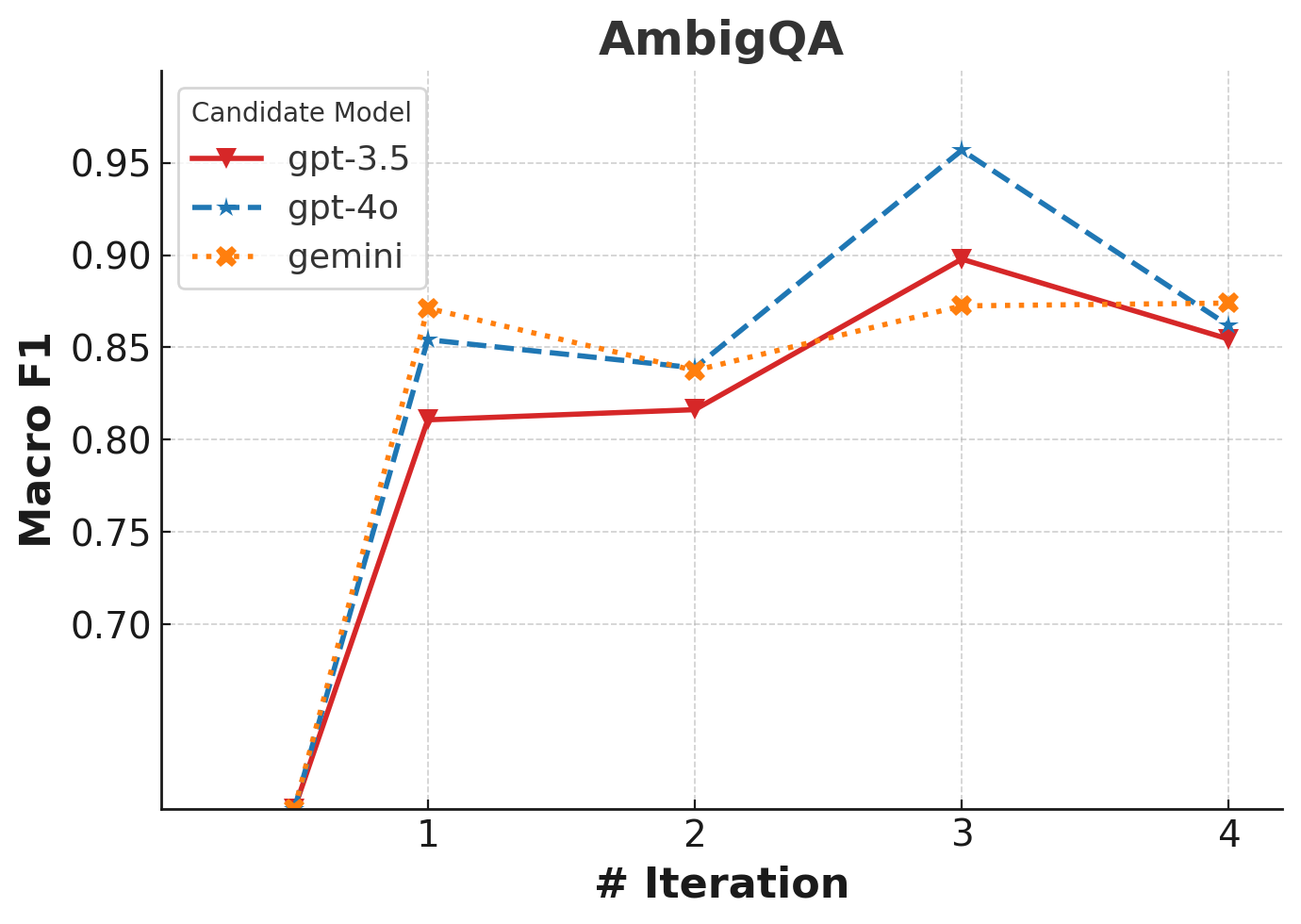}}
\end{minipage}
\hspace{0.02\textwidth}
\begin{minipage}{0.40\textwidth}
  \fbox{\includegraphics[width=\linewidth,keepaspectratio]{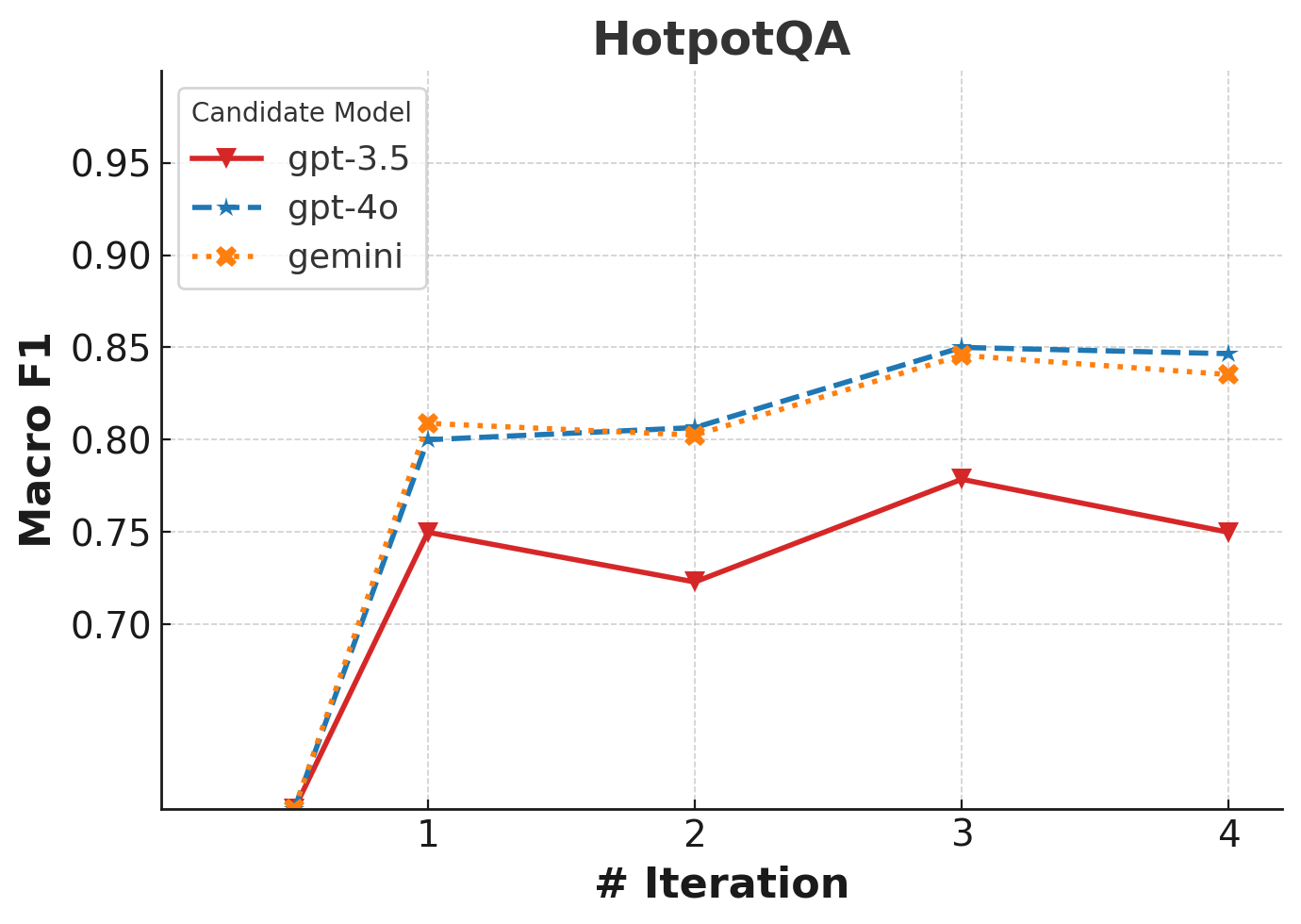}}
\end{minipage}
\end{center}
\caption{The effect of iterations. GPT-4o is used as a judge here.}
\label{fig:iteration_effect_side_by_side}
\end{figure*}

\paragraph{Effect of iterations.}
Figure~\ref{fig:iteration_effect_side_by_side} shows the effect of the number of SAGE iterations on judge performance. Any positive number of iterations improves over the zero-iteration (parametric-only) baseline, and three iterations---our default---consistently offers the best trade-off between performance and cost. A small dip at two iterations reflects occasional off-topic refinement queries that are typically corrected in the following round, while the decline at four iterations is driven by an overabundance of sources that inflates context length and introduces redundant or irrelevant evidence.

\paragraph{Effect of query generation.}
To evaluate the contribution of the query-generation module, we remove it and use the input question directly for evidence retrieval. Table~\ref{tab:query_comparison_gpt4o} shows that SAGE consistently outperforms this query-free baseline, demonstrating the importance of the query-generation step. The most notable improvements are on AmbigQA, where SAGE reaches a Macro-F1 of 0.96 compared to 0.84 without query generation. On HotpotQA the gain is smaller but still positive, reflecting SAGE's ability to adaptively generate focused queries that facilitate multi-hop reasoning.

\begin{table}[t]
\centering
\footnotesize
\begin{tabular}{llcc}
\toprule
\textbf{Candidate} & \textbf{Task} & \textbf{Judge w/o Query} & \textbf{SAGE} \\
\midrule
\textbf{GPT-3.5} & AmbigQA  & 0.84 & 0.90 \\
                 & HotpotQA & 0.76 & 0.78 \\
\midrule
\textbf{GPT-4o}  & AmbigQA  & 0.84 & 0.96 \\
                 & HotpotQA & 0.82 & 0.85 \\
\midrule
\textbf{Gemini}  & AmbigQA  & 0.85 & 0.87 \\
                 & HotpotQA & 0.80 & 0.85 \\
\bottomrule
\end{tabular}
\caption{Macro F1 scores between Judge w/o Query and SAGE using GPT-4o as the judge. Note that directly using the input question without generating refined queries is still considered a form of querying, but for clarity, we refer to this setting as w/o Query.}
\label{tab:query_comparison_gpt4o}
\end{table}

\paragraph{Robustness of query refinement.}
We examined whether SAGE’s query refinement introduces topic drift or irrelevant queries by manually analyzing 100 instances from our error analysis. Since our SAGE configuration involves up to three iterations, this requires analyzing the queries at each stage. One 3rd-iteration query was excluded because annotators could not reach consensus on its category, leaving 299 queries in total. Table~\ref{tab:quries_topic_drift} shows that initial queries generated directly from the input question are generally on-topic and relevant to the core question being asked. However, due to the ambiguity of some questions (e.g., AmbigQA), we observe some off-topic queries in this stage. Topic drift, though not the most dominant error, did occur during the refinement stages (iterations 2 and 3). Out of the 199 refined queries analyzed in the refinement stages, only 6.5\% of queries were judged ``insufficient.''

\begin{table}[t]
\centering
\footnotesize
\setlength{\tabcolsep}{3.5pt}
\begin{tabular}{lccc}
\toprule
\textbf{Category} & \textbf{Init. Query} & \textbf{2. Query} & \textbf{3. Query} \\
\midrule
Sufficient (On-topic)        & 77  & 69 & 70 \\
Partial sufficient           & 15  & 25 & 22 \\
Insufficient (Off-topic)
   & 8   & 6  & 7  \\
\bottomrule
\end{tabular}
\caption{Topic drift across 300 queries (100 inst $\times$ 3).}
\label{tab:quries_topic_drift}
\end{table}

\paragraph{Impact of iterative evidence gathering.}
To isolate the value of iterative retrieval, we compare against a minimal search-augmented baseline that performs a single-pass retrieval: it issues one web query from the input question, collects the top-3 snippets, and feeds them to the judge without summarization, reflection, or refinement. We write \textbf{SAGE-$k$} for SAGE restricted to $k$ iterations; our default configuration is SAGE-3. All configurations in this ablation use GPT-4o as both candidate and judge.

On AmbigQA, the single-pass baseline already raises the judge-without-search $\kappa$ from 0.38 to 0.65, confirming that one round of external evidence yields a substantial improvement over parametric-only judging. Restoring SAGE's summarize--reflect step under the same one-query budget (SAGE-1) lifts $\kappa$ further to 0.74 ($+0.09$), because the reflect step explicitly tests evidence sufficiency and rejects false-positive acceptances when the top-$k$ is off-topic. Running the full three-iteration loop (SAGE-3) pushes $\kappa$ to 0.91. HotpotQA shows the same trend: $\kappa$ climbs from 0.38 (no search) to 0.58 (single-pass), 0.64 (SAGE-1), and 0.70 (SAGE-3). These results isolate two distinct sources of gain: external evidence \emph{per se}, and the reflect-and-refine loop that turns that evidence into a reliable verdict.

\paragraph{Robustness to the search engine.}
Because SAGE depends on an external retriever, we ask whether its gains persist when the search engine is replaced. We rerun the full pipeline with Brave and Tavily in place of Serper, keeping all other components fixed. Table~\ref{tab:search_engine_robustness} reports Cohen's $\kappa$ for the single-pass baseline, SAGE-1, and SAGE-3 on AmbigQA and HotpotQA with GPT-4o as candidate and judge. The iterative gain from SAGE is preserved across all three engines, and SAGE-3 remains the strongest configuration in every case. 

\begin{table}[t]
\centering
\footnotesize
\setlength{\tabcolsep}{3pt}
\begin{tabular}{llccc}
\toprule
\textbf{Task} & \textbf{Engine} & \textbf{1-pass} & \textbf{SAGE-1} & \textbf{SAGE-3} \\
\midrule
\multirow{3}{*}{\textbf{AmbigQA}}
  & Serper & 0.65 & 0.74 & \textbf{0.91} \\
  & Brave  & 0.60 & 0.72 & 0.85 \\
  & Tavily & 0.45 & 0.60 & 0.70 \\
\midrule
\multirow{3}{*}{\textbf{HotpotQA}}
  & Serper & 0.58 & 0.64 & \textbf{0.70} \\
  & Brave  & 0.60 & 0.66 & \textbf{0.71} \\
  & Tavily & 0.36 & 0.62 & 0.67 \\
\bottomrule
\end{tabular}
\caption{Robustness of SAGE to the search engine. Cohen's $\kappa$ against the human majority on AmbigQA and HotpotQA, with GPT-4o as both candidate and judge.
}
\label{tab:search_engine_robustness}
\end{table}

\subsection{Cost and latency}
As given in Table~\ref{tab:search_engine_robustness}, a full SAGE-3 evaluation makes 10 LLM calls and 3 web-search queries per instance. At April~2026 list prices for GPT-4o-mini\footnote{\$0.15 / \$0.60 per $10^{6}$ input / output tokens; \url{https://openai.com/api/pricing/}.} and Serper's Starter plan,\footnote{\$1.00 per $10^{3}$ queries; \url{https://serper.dev/}.} this works out to about $\$0.0043$ per instance, so the full 300-instance AmbigQA subset is evaluated end-to-end for roughly $\$1.28$---orders of magnitude cheaper than manual annotation. Client-side wall-clock, measured on a 50-instance sample (Table~\ref{tab:latency}), averages $\approx 27$\,s per instance, of which $\approx 23$\,s is LLM time spread across 10 sequential calls (network round-trip, queueing, and JSON handling) and only $\approx 3$\,s is search. Since instances are independent, the batch parallelizes trivially: a 300-item dataset can be evaluated in minutes with modest concurrency.

\begin{table}[t]
\centering
\scriptsize
\setlength{\tabcolsep}{3pt}
\begin{tabular}{lcccc}
\toprule
\textbf{Setting} & \textbf{Total (s)} & \textbf{LLM (s)} & \textbf{Search (s)} & \textbf{\#LLM} \\
\midrule
LLM-as-Judge & $1.65_{\pm0.98}$  & ---              & ---              & 1  \\
1-pass       & $3.15_{\pm1.24}$  & $2.18_{\pm1.17}$ & $0.96_{\pm0.40}$ & 1  \\
SAGE-3       & $26.32_{\pm3.10}$ & $23.16_{\pm3.06}$ & $3.13_{\pm0.59}$ & 10 \\
\bottomrule
\end{tabular}
\caption{End-to-end latency per instance on 50 AmbigQA items (GPT-4o-mini at $T{=}0$ via Serper). Values are mean\,$_{\pm\mathrm{std}}$. LLM time dominates SAGE-3; the three search calls together contribute only $\sim$3\,s.}
\label{tab:latency}
\end{table}
\section{Related Work}
\label{sec:related_work}
Evaluating LLMs is a critical yet challenging aspect of modern NLP research. We review
existing approaches across the following categories.

\paragraph{Free-form QA.} It is a valuable benchmark for ensuring the factuality of LLMs~\citep{wang2023evaluating}. This type of task is traditionally evaluated through automatic metrics that rely on comparing model outputs against expert-annotated reference answers using metrics such as EM and F1~\citep{gou2024criticllms}. While efficient, such methods cannot capture the diversity of responses, require costly reference annotations, and fail to adapt to evolving factual information~\citep{kamalloo-etal-2023-evaluating}.

\paragraph{Reference-based LLM judge.} Recent work has attempted to address such limitations by utilizing LLMs for evaluation. Specifically, given the model answer for a question, an LLM is prompted with the original question, the candidate answer, and the dataset reference answer to evaluate the correctness of the model response~\citep{wang2023evaluating, kamalloo-etal-2023-evaluating}. This approach often returns a verdict in the form of a categorical label or a scalar score. Recent methods, for instance, PoLL~\citep{verga2024replacingjudgesjuriesevaluating} follow a similar template but utilize multiple LLMs for more reliable evaluations.

\paragraph{Reference-free LLM judge.} To avoid the need for reference answers to improve scalability, subsequent work explored reference-free LLM judges~\citep{zheng2023judging}. G-Eval~\citep{liu2023g} implements direct evaluation by prompting models to assess outputs based on predefined criteria. Other methods include pairwise comparisons~\citep{zheng2023judging}, debate-style frameworks~\citep{khan2024debating}, and ensemble approaches~\citep{zhang2024large}. These methods have demonstrated success in subjective evaluation tasks such as summarization or dialogue generation, where human preferences rather than factual correctness are the primary concern. However, for objective correctness, reference-free LLM judges often struggle with reliability~\citep{badshah-etal-2025-clev, DAFE, kim2024prometheus2opensource}, because although we can provide them with detailed instructions at inference time, their factual grounding still depends entirely on the parametric knowledge encoded in their pre-trained weights and thus inherits its limitations.
 
\paragraph{LLMs evaluation with search-augmentation.} An emerging category attempts to overcome the limitations of LLM evaluators by incorporating external tools. More closely related to our work, FActScore~\citep{min2023factscore} decomposes long-form generated text into atomic facts and verifies each against Wikipedia pages. Similarly, SAFE~\citep{wei2024long} uses an LLM to split long-form responses into individual facts, issue a Google Search query for each, and reason about relevance. In contrast to these methods, which target long-form outputs and rely on splitting text into atomic facts, our focus is short-form responses, where the challenge is less about exhaustive claim coverage and more about precise, reference-free evaluation under uncertainty. Unlike prior search-augmented evaluators that typically issue a single-pass query, our method iteratively conducts output-specific searching, summarization, reflection, and refinement. This added iteration increases cost relative to a single pass but provides greater reliability, particularly in ambiguous cases.

\section{Conclusion}
We presented SAGE, a reference-free framework that equips an LLM judge with iterative web retrieval, evidence summarization, reflection, and query refinement to evaluate the factual correctness of free-form QA responses. Unlike reference-based metrics that require costly pre-annotated answers, and unlike parametric-only judges that are prone to hallucinated rationales, outdated knowledge, and uncritical acceptance of candidate claims, SAGE grounds every verdict in externally retrieved, verifiable evidence.

Experiments across five QA benchmarks and different model families demonstrate that SAGE achieves substantial to perfect agreement with human evaluators, consistently outperforming both lexical baselines and reference-free judges. Ablations show that the gains arise from two complementary sources: external evidence retrieval itself, and the iterative summarize--reflect--refine loop that converts raw snippets into reliable verdicts. These benefits hold across multiple search engines, prompt designs, and smaller open-weight judges, while remaining orders of magnitude cheaper than human annotation.

Looking ahead, we see two natural extensions. First, adaptive stopping based on reflection-derived sufficiency signals can reduce latency on straightforward instances without sacrificing reliability on harder ones. Second, SAGE's summarize--reflect--refine loop is domain-agnostic; by replacing the web-search tool with a code interpreter, a calculator, or a knowledge-base API, the same framework could support evaluation of code generation, mathematical reasoning, and structured-knowledge tasks.
\section{Limitations}
\label{sec:limitations}

\paragraph{Context window constraints.} SAGE's short-term memory grows with each iteration as (query, evidence, reflection) tuples accumulate, so the judge's context window bounds how many iterations can be chained before relevant earlier steps are truncated. Future work could replace the append-only buffer with a recall-based long-term memory that selectively retrieves past traces. For example, episodic and semantic memory~\citep{park2023generativeagentsinteractivesimulacra} to reduce both context pressure and computational cost.

\paragraph{Source bias and quality control.} SAGE's reflection module detects inconsistencies across retrieved sources but does not explicitly model source credibility. When multiple sources agree on an incorrect or outdated claim, the judge has no mechanism to discount them, and the evaluation may inherit biases from the external data~\citep{li2025commercialllmagentsvulnerable, zhan2024injecagentbenchmarkingindirectprompt, yu-etal-2024-latent}. Integrating credibility scoring or provenance-aware weighting that dynamically assess evidence reliability is a natural extension.

\paragraph{Dependency on judge-LLM capabilities.} SAGE's performance depends on the capabilities of the underlying judge model. Our experiments show a noticeable drop in performance with smaller judges such as Mistral~7B (see Appendix~\ref{app:small_llm}). Deploying SAGE with very small judges therefore involves a reliability--cost trade-off that users should weigh explicitly.

\paragraph{Diminishing returns with iterative refinement.} Performance gains plateau and occasionally degrade beyond three iterations as redundant or off-topic evidence accumulates and the context window fills up; we therefore fix $N = 3$ empirically. In future work, we plan to explore adaptive stopping: a lightweight heuristic could use the existing reflection signal (evidence sufficiency combined with low evidence novelty across consecutive iterations) to terminate early, while more principled variants such as instruction-tuning the reflection module to output a calibrated sufficiency score, or conformal-prediction-based stopping~when the estimated flip-risk falls below a target level could further improve the efficiency--reliability trade-off.

\paragraph{Domain scope.} SAGE is validated on factual free-form QA, where claims can be verified against web-retrievable evidence. Extending SAGE to other evaluation domains such as code generation (via an interpreter or unit tests), mathematical reasoning (via a calculator or computer algebra system), or structured knowledge evaluation, would require swapping the web-search tool for domain-appropriate tools while retaining the summarize--reflect--refine loop. We leave this generalization to future work.

\paragraph{Binary verdicts.} SAGE outputs a binary True/False verdict, which is well-suited to factual-correctness evaluation but does not support partial-correctness scoring, pairwise comparison, or rubric-based ratings. Tasks such as summarization quality or instruction-following preference, where judgments are inherently graded, would require extending SAGE's verdict schema beyond binary classification.

\paragraph{Language coverage.} All experiments in this paper are conducted in English. SAGE's effectiveness on non-English QA where web-search quality, snippet relevance, and the judge LLM's multilingual reasoning capabilities all vary---remains untested.

\section*{Acknowledgment}
We acknowledge the support of the Natural Sciences and Engineering Research Council of Canada (NSERC), Canada Foundation for Innovation (CFI), and Research Nova Scotia. Advanced computing resources are provided by ACENET, the regional partner in Atlantic Canada, and the Digital Research Alliance of Canada.

\bibliography{custom}
\section{Experimental detail}
\subsection{Datasets}\label{app:datasets}
We evaluate SAGE on widely used free-form question-answering datasets that span different question types, knowledge domains, and complexity levels. Evaluating large-scale datasets is costly, so we randomly sample 300 instances per dataset. Each dataset's reference answers serve as ground truth for our reference-based baselines. Our selected datasets are:

\paragraph{AmbigQA}~\citep{minetal2020ambigqa} Contains questions with multiple valid answers due to inherent ambiguities, challenging evaluators to consider multiple interpretations.

\paragraph{HotpotQA}~\citep{yang2018hotpotqa} Features multi-hop reasoning questions that require synthesizing information from multiple sources.

\paragraph{Natural Questions (NQ-Open)}~\citep{kwiatkowskietal2019} Consists of real user queries from Google Search, representing naturally occurring information needs.

\paragraph{TriviaQA}~\citep{joshi2017triviaqa} Includes trivia questions from various domains, testing breadth of knowledge and factual recall.

\paragraph{FreshQA}~\citep{vu2023freshllms} Contains questions about recent events occurring after most LLMs' training cutoff, specifically designed to test knowledge updating capabilities.

\noindent We use the respective validation splits of each dataset: the standard validation sets for AmbigQA and Natural Questions, the \texttt{distractor} subset validation split for HotpotQA, and the \texttt{unfiltered.nocontext} validation subset for TriviaQA. For FreshQA, we adopt the version released on December 18, 2024.

\subsection{Prompting}\label{app:prompting}
We employ a template-based prompting strategy for both response generation and evaluation. For candidate models, we utilize few-shot Chain-of-Thought (CoT) prompts~\citep{gou2024criticllms}, incorporating 6 examples per dataset to encourage detailed, reasoning-driven, and structured responses (see Figure~\ref{fig:answer_generation_example}).

\begin{figure*}[t]
\centering
\begin{tcolorbox}[colback=gray!5, colframe=gray!60, title=\textbf{Candidate Answer Generation (Few-shot CoT)}, sharp corners=south, width=0.95\linewidth, boxrule=0.4pt]
\small
\textbf{Instructions:}  
Answer each question step by step and conclude with: “So the answer is: [Answer].”

\vspace{2pt}
\textbf{Q:} What is the elevation range for the area that the eastern sector of the Colorado orogeny extends into?  
\textbf{A:} The eastern sector of the Colorado orogeny extends into the High Plains, which rise from around 1,800 to 7,000 ft.  
\textbf{So the answer is:} 1,800–7,000 ft.

\vspace{2pt}
\textbf{Q:} Musician and satirist Allie Goertz wrote a song about the "The Simpsons" character Milhouse, who Matt Groening named after who?  
\textbf{A:} Milhouse was named after U.S. president Richard Nixon.  
\textbf{So the answer is:} Richard Nixon.

\vspace{2pt}
\textbf{Q:} Which documentary is about Finnish rock groups, \textit{Adam Clayton Powell} or \textit{The Saimaa Gesture}?  
\textbf{A:} \textit{Adam Clayton Powell} is about an African-American politician, so the documentary about Finnish rock groups must be \textit{The Saimaa Gesture}.  
\textbf{So the answer is:} The Saimaa Gesture.

\vspace{2pt}
\textbf{Q:} What profession do Nicholas Ray and Elia Kazan have in common?  
\textbf{A:} Both were directors, screenwriters, and actors.  
\textbf{So the answer is:} director, screenwriter, actor.

\vspace{2pt}
\textbf{Q:} Which magazine was started first, \textit{Arthur's Magazine} or \textit{First for Women}?  
\textbf{A:} \textit{Arthur's Magazine} (1844) was started before \textit{First for Women} (1989).  
\textbf{So the answer is:} Arthur's Magazine.

\vspace{2pt}
\textbf{Q:} Were Pavel Urysohn and Leonid Levin known for the same type of work?  
\textbf{A:} Both were mathematicians (Levin is also a computer scientist).  
\textbf{So the answer is:} Yes.
\end{tcolorbox}
\caption{Examples of few-shot CoT~\citep{gou2024criticllms} prompts for candidate answer generation.}
\label{fig:answer_generation_example}
\end{figure*}

The \emph{candidate} model generates responses under the few-shot CoT template above. The \emph{SAGE} modules use a separate, one-shot role prompt per module that combines role-playing instructions with explicit reasoning goals; these prompts are shared across datasets and judges. Below, we describe the prompting strategy for each SAGE component.

\paragraph{Query Generation.}
The query generation module converts an input question $x$ into an initial search query without referencing the candidate answer. The prompt instructs the model to reflect step-by-step on the most relevant aspects and keywords before proposing a final query.

\begin{figure*}[t]
\centering
\begin{tcolorbox}[colback=gray!5, colframe=gray!60, title=\textbf{Query generation}, sharp corners=south, width=0.95\linewidth, boxrule=0.4pt]
\small
\textbf{Your goal is to generate a targeted web search query.}

Before producing the final query, carefully consider:
\begin{enumerate}
    \item The question's key concepts or keywords (e.g., important names, dates).
    \item Whether the question might be ambiguous or reference multiple possible answers (e.g., a book with the same title by different authors, or a modern text about a historical figure).
\end{enumerate}

\vspace{3pt}
\textbf{Question:} \{question\}

\vspace{3pt}
\textbf{Return your response as a JSON object with ALL three exact keys:}
\begin{itemize}
    \item \texttt{"query"}: The search query string.
    \item \texttt{"aspect"}: The specific aspect of the question to focus on.
    \item \texttt{"rationale"}: A brief explanation of why this query is relevant, including your chain-of-thought reasoning.
\end{itemize}

\vspace{3pt}
\textbf{Example Output:}
\begin{verbatim}
{
  "query": "Apollo 11 moon landing year + NASA + 1969",
  "aspect": "historical event",
  "rationale": "The question asks about Apollo 11's landing year, 
                so I'm including NASA, year, and 1969 to get relevant info."
}
\end{verbatim}
\end{tcolorbox}
\caption{Prompt used for initial query generation, guiding the model to produce focused and relevant search queries.}
\vspace{-5pt}
\label{fig:query_generation_prompt}
\end{figure*}

\paragraph{Evidence Summarization.}
To reduce raw search results \( S(q_i) \), the summarization module uses a CoT prompt that walks the model through evaluating and synthesizing relevant content. The prompt emphasizes factual grounding and asks the model to avoid repetition and speculation. 

\begin{figure*}[t]
\centering
\begin{tcolorbox}[colback=gray!5, colframe=gray!60, title=\textbf{Evidence summarization}, sharp corners=south, width=0.95\linewidth, boxrule=0.4pt]
\small
\textbf{You are a summarization assistant. Carefully review the raw search results and provide a concise summary of the key information relevant to the question.}

\vspace{3pt}
\textbf{Raw Search Results:} \{raw\_results\}

\vspace{3pt}
\textbf{Return your summary as plain text:}
\begin{itemize}
    \item Keep it neutral and focused on the question.
    \item If results conflict, mention that briefly.
    \item Do not add extra commentary.
\end{itemize}

\vspace{3pt}
\textbf{Example Output (Plain Text):}
\begin{verbatim}
"Result 1 says X about the event date, 
 Result 2 says Y but doesn't mention the exact date. 
 Overall, it references 1969."
\end{verbatim}
\end{tcolorbox}
\caption{Prompt for evidence summarization, guiding the model to generate a concise, unbiased summary from raw search results.}
\vspace{-5pt}
\label{fig:evidence_summarization_prompt}
\end{figure*}

\paragraph{Iterative Reflection.}
The reflection module analyzes the current evidence summary \( E_i \) in relation to the input question \( x \) and candidate answer \( \hat{y} \). The prompt guides the model to assess whether the evidence supports, contradicts, or is inconclusive with respect to the answer, and highlights missing information. 

\begin{figure*}[t]
\centering
\begin{tcolorbox}[colback=gray!5, colframe=gray!60, title=\textbf{Example prompt for iterative reflection}, sharp corners=south, width=0.95\linewidth, boxrule=0.4pt]
\small
\textbf{You are a research assistant tasked with analyzing the gathered evidence in relation to the question and candidate answer. Think step by step—explain your reasoning and note any gaps or additional details that might be needed. Do not provide a final decision; simply offer your chain-of-thought reflection.}

\vspace{3pt}
\textbf{Question:} \{question\} \\
\textbf{Candidate Answer:} \{candidate\_answer\} \\
\textbf{Evidence Summary:} \{evidence\_summary\}

\vspace{3pt}
\textbf{Return your response as a JSON object with a single key:}
\begin{itemize}
    \item \texttt{"reflection"}: Your chain-of-thought reflection summarizing your analysis.
\end{itemize}

\vspace{3pt}
\textbf{Example Output:}
\begin{verbatim}
{
  "reflection": "I observed that the evidence overwhelmingly confirms
                 that Apollo 11 landed on the moon in 1969, though there 
                 is slight variation in the reported landing times across 
                 sources. Additional authoritative sources might help 
                 resolve these minor discrepancies."
}
\end{verbatim}
\end{tcolorbox}
\caption{Prompt for iterative reflection, instructing the model to analyze the relationship between the question, candidate answer, and evidence summary.}
\vspace{-5pt}
\label{fig:iterative_reflection_prompt}
\end{figure*}

\paragraph{Query Refinement.}
To improve retrieval in subsequent iterations ($i > 1$), the query refinement prompt conditions on the accumulated short-term memory $\mathcal{M}_{i-1} = \bigl((q_1, E_1, r_1), \ldots, (q_{i-1}, E_{i-1}, r_{i-1})\bigr)$ and produces a new query $q_i$. Concretely, the prompt receives the question $x$, the most recent query, an aggregated evidence summary over all prior iterations, and the latest reflection, and is instructed via CoT to identify remaining uncertainties or gaps before generating a refined, more targeted query.

\begin{figure*}[t]
\centering
\begin{tcolorbox}[colback=gray!5, colframe=gray!60, title=\textbf{Example prompt for query refinement}, sharp corners=south, width=0.95\linewidth, boxrule=0.4pt]
\small
\textbf{You are a research assistant. Before refining the search query, analyze the existing evidence and reflect on what keywords might be missing or need emphasis. Think step by step and then produce your final refined query.}

\vspace{4pt}
\textbf{Question:} \{question\} \\
\textbf{Current Search Query:} \{current\_query\} \\
\textbf{Aggregated Evidence Summary:} \{evidence\_summary\} \\
\textbf{Iterative Reflection:} \{iterative\_reflection\}

\vspace{4pt}
If the evidence still does not resolve the question or if there might be an alternative perspective, incorporate additional, more specific keywords to explore those possibilities. For instance:
\begin{itemize}
    \item Add relevant dates or historical context.
    \item Use synonyms or alternate phrasings for ambiguous or repeated terms.
    \item Specify a domain or subject area (e.g., “film,” “novel,” “historical figure”) if it reduces confusion.
    \item Highlight the location, time period, or any unique aspect not yet included in the current query.
\end{itemize}

\vspace{4pt}
\textbf{Return your response as a JSON object with ALL three exact keys:}
\begin{itemize}
    \item \texttt{"query"}: The refined search query.
    \item \texttt{"aspect"}: The specific aspect being targeted with the refined query.
    \item \texttt{"rationale"}: A brief explanation of your reasoning (chain-of-thought) and why this refinement is needed.
\end{itemize}

\vspace{4pt}
\textbf{Example Output:}
\begin{verbatim}
{
  "query": "Apollo 11 detailed timeline moon landing 1969",
  "aspect": "chronological sequence",
  "rationale": "The initial query did not specify the temporal progression
                of events. I refined it to target a detailed timeline of
                the Apollo 11 mission in 1969 to capture the sequence of
                key events."
}
\end{verbatim}
\end{tcolorbox}
\caption{Prompt for query refinement, guiding the model to analyze evidence and generate more targeted queries.}
\vspace{-5pt}
\label{fig:query_refinement_prompt}
\end{figure*}

\paragraph{Judgment.}
Finally, the judgment module implements $J(x, \hat{y}, \mathcal{M}_N)$: given the question, the candidate answer, and the accumulated memory $\mathcal{M}_N$, it emits a binary verdict $v \in \{0, 1\}$ together with a natural-language rationale $\rho$. The CoT prompt conditions on the question, the candidate answer, the aggregated evidence summary, and the aggregated reflection string---all materialized from $\mathcal{M}_N$---and instructs the model to reason step by step before committing to its verdict. The output is formatted as a JSON object with keys \texttt{``decision''} and \texttt{``explanation''}, corresponding to $v$ and $\rho$ respectively.

\begin{figure*}[t]
\centering
\begin{tcolorbox}[colback=gray!5, colframe=gray!60, title=\textbf{Prompt to the judge model}, sharp corners=south, width=0.95\linewidth, boxrule=0.4pt]
\small
\textbf{You are a critical evaluator. You have:}\\
1. The question and the candidate answer, \\
2. The evidence summary from multiple iterative searches (which may contain overlapping or conflicting information), \\
3. The chain-of-thought reflection from prior steps, \\
4. Your own broad knowledge (only if the above are inconclusive).\\[3pt]

\textbf{Follow these guidelines:}\\
- If the summarized evidence and reflections strongly conflict with the candidate answer, conclude \texttt{"False"}. \\
- If the evidence strongly confirms the candidate answer, conclude \texttt{"True"}. \\
- If the evidence is inconclusive or incomplete, but your own knowledge supports the answer, you may conclude \texttt{"True"} if confident. Otherwise, conclude \texttt{"False"} or state insufficient information. \\
- When the retrieved evidence is irrelevant, prioritize the chain-of-thought reflections and your own knowledge.\\[3pt]

\textbf{Produce your conclusion in JSON with:}\\
- \texttt{"decision"}: \texttt{"True"} or \texttt{"False"} \\
- \texttt{"explanation"}: A concise reason (including your step-by-step reasoning) describing how you arrived at the verdict.\\[3pt]

\textbf{Input:}\\
\textbf{Question:} \{question\} \\
\textbf{Candidate Answer:} \{candidate\_answer\} \\
\textbf{Evidence Summary:} \{evidence\_summary\} \\
\textbf{Reflection:} \{reflection\}\\[3pt]

\textbf{Example Output:}
\begin{verbatim}
{
  "decision": "True",
  "explanation": "The evidence overwhelmingly confirms that Apollo 11
                  landed on the moon in 1969. While minor discrepancies
                  exist in the reported times, they do not undermine the
                  main conclusion. Additional verification is unnecessary."
}
\end{verbatim}
\end{tcolorbox}
\caption{Prompt for the judgment step, instructing the model to analyze evidence and reflections to generate a final verdict with justification.}
\vspace{-5pt}
\label{fig:judgment_prompt}
\end{figure*}

\subsection{Baselines}
\label{app:baselines}
We compare SAGE against several established evaluation approaches:

\subsubsection{Reference-based metrics} We implement three reference-based baselines that rely on comparison with dataset-specific reference answers:
\begin{itemize}
    \item \textbf{Exact Match (EM)} measures whether the model's answer exactly matches any of the reference answers after normalization.
    \item \textbf{F1 Score} computes the harmonic mean of precision and recall between the token sets of the model's answer and the references, providing a softer measure of overlap.
    \item \textbf{RefGPT} prompts GPT-4 with the question, candidate answer, and dataset reference answers, and asks it to issue a binary correctness verdict~\citep{wang2023evaluating}. This provides a context-aware evaluation beyond strict token-level matching.
\end{itemize}

\subsubsection{Judge without search} Following~\citet{liu2023g}, we implement a reference-free baseline in which the judge LLM evaluates candidate answers based solely on the question--answer pair, without access to external tools or reference answers. The judge relies entirely on its pre-trained knowledge to determine factual correctness. This baseline isolates the impact of search augmentation in SAGE by holding the judge model fixed and removing only the evidence-retrieval mechanism.

\subsubsection{Human evaluation}\label{app:human_eval}
The main setup including annotators, randomization, and the binary 1/0 scoring scheme is described in \S\ref{sec:experiments}. Here we add the budgetary rationale and the full annotator guidelines.

\paragraph{Scale and rationale.}
We limit human evaluation to AmbigQA and HotpotQA because these datasets best exercise SAGE's core use case (ambiguous or multi-hop questions) and because extending to all five datasets would have significantly increased annotation cost. For each of the two tasks we evaluate 300 instances per candidate model, and we evaluate three candidate models, for a total of $2 \times 300 \times 3 = 1{,}800$ annotated judgments.

\paragraph{Evaluation Guidelines}
To ensure consistent assessments, annotators followed the guidelines inspired by established evaluation protocols. Annotators were instructed to evaluate responses based on the following principles:

\begin{itemize}
    \item \textbf{Semantic equivalence:} A response is marked \textbf{True} if it conveys the same core information as the reference answer, even if phrased differently using synonyms, paraphrasing, or structural variations. Additional contextual information is acceptable as long as it is factually correct and does not alter the original meaning.
    
    \item \textbf{Factual Accuracy:} Responses that contain factual errors, omit essential information, or introduce misleading content are marked \textbf{False}. If a response partially answers the question but excludes critical elements, it is considered incorrect.
    
    \item \textbf{Multiple Reference Answers:} In cases with multiple reference answers, a response is deemed correct if it is fully aligned with at least one reference.
    
    \item \textbf{Fact-Checking:} Annotators are allowed to consult external resources, such as search engines or online encyclopedias, to verify specific facts when uncertain. However, the reference answers served as the primary benchmark for correctness.
    
    \item \textbf{Documenting Ambiguity:} Annotators are encouraged to document cases where the evaluation is uncertain or requires further clarification. These cases were discussed collaboratively to ensure consensus.
\end{itemize}

By adhering to these guidelines, we ensured reliable and consistent human evaluations.

\paragraph{Inter Human Annotator Agreement}
We calculated \textbf{Fleiss’ Kappa ($\kappa$)} and percent agreement to measure inter-annotator agreement. Fleiss’ Kappa is defined as:
\[
\kappa = \frac{\bar{P} - P_e}{1 - P_e},
\]
where $\bar{P}$ is the average observed agreement among annotators, and $P_e$ is the expected agreement by chance. Percent agreement (PA) is calculated as:
\[
\text{PA} = \frac{\text{N.Agreements}}{\text{Total N.Annotations}} \times 100.
\]

\subsection{Evaluation Metrics}
To assess SAGE's performance, we use multiple evaluation metrics:

\paragraph{Accuracy:} We measure the proportion of instances where the judge's binary verdict (correct/incorrect) matches the reference label obtained from automatic metrics or from RefGPT.

\paragraph{Agreement with Human Judgment:} For the AmbigQA and HotpotQA subsets with human annotations, we calculate Cohen's Kappa ($\kappa$), majority voting, and Macro-F1 scores to assess agreement between SAGE's verdicts and human majority votes. These metrics were chosen because they account for both agreement beyond chance ($\kappa$) and class balance (Macro-F1). \\

\noindent\textbf{Cohen's Kappa:} Cohen's Kappa measures the agreement between two annotators while correcting for chance agreement. It is defined as:

\[
\kappa = \frac{P_o - P_e}{1 - P_e},
\]

\noindent where \( P_o \) is the observed agreement, and \( P_e \) is the expected agreement by chance. \\

\noindent\textbf{Majority Voting:} In majority voting, the final decision is determined based on the majority of annotators' labels. Given \( n \) annotators and a binary classification, the majority label is defined as:

\[
y_{\text{majority}} = \begin{cases} 
1 & \text{if } \sum_{i=1}^{n} y_i > \frac{n}{2}, \\
0 & \text{otherwise},
\end{cases}
\]

\noindent where \( y_i \) represents the label assigned by the \( i \)th annotator.

\noindent\textbf{Macro F1 Score:} Macro F1 evaluates the balance between precision and recall for each class and averages the results. It is calculated as:

\[
\text{Macro-F1} = \frac{1}{C} \sum_{c=1}^{C} \frac{2 \cdot \text{Precision}_c \cdot \text{Recall}_c}{\text{Precision}_c + \text{Recall}_c},
\]

\noindent where \( C \) is the number of classes, and \( \text{Precision}_c \) and \( \text{Recall}_c \) are the precision and recall for class \( c \).

\section{Additional results}\label{app:additional_results}
In this section, we included additional results obtained through our experiments.

\subsection{Inter-human annotator agreement}\label{app:inter_human_annotator_agreement}

Table \ref{tab:human_annotator_agreement} presents the human annotator agreement results for the AmbigQA and HotpotQA across three candidate models. The results indicate consistently high agreement among annotators.

\begin{table}[t]
\centering
\resizebox{\linewidth}{!}{
\begin{tabular}{l l c c c}
\toprule
\textbf{Task} & \textbf{Model} & \textbf{Percent Agreement (\%)} & \textbf{Fleiss' Kappa} & \textbf{Samples} \\
\midrule
\multirow{3}{*}{AmbigQA}
 & GPT-3.5  & 98.3 & 0.972 & 300 \\
 & GPT-4o   & 98.3 & 0.976 & 300 \\
 & Gemini   & 97.0 & 0.953 & 300 \\
\midrule
\multirow{3}{*}{HotpotQA}
 & GPT-3.5  & 98.3 & 0.978 & 300 \\
 & GPT-4o   & 98.3 & 0.978 & 300 \\
 & Gemini   & 98.3 & 0.977 & 300 \\
\bottomrule
\end{tabular}
}
\caption{Human annotator agreement results on AmbigQA and HotpotQA tasks.}
\vspace{-5pt}
\label{tab:human_annotator_agreement}
\end{table}

\subsection{SAGE agreement with reference-based metrics}
A reference-free judge is only useful if its verdicts track reference-based signals when the latter are reliable. Table~\ref{tab:full_metrics} reports the per-evaluator accuracy for every (candidate, task) pair, alongside the lexical EM/F1 baselines and RefGPT. Two patterns stand out. First, SAGE's raw accuracy tracks the reference-based signals closely: for example, GPT-3.5 as a SAGE judge reaches 0.64 when evaluating its own answers on AmbigQA, essentially matching its reference-based F1 of 0.63. Second, judges without search exhibit a consistent self-enhancement bias---e.g., GPT-3.5 inflates its own AmbigQA accuracy to 0.81 when asked to judge without evidence, which is absent under SAGE. EM, by contrast, often under-estimates model quality because it misses valid paraphrases and alternative formulations. SAGE therefore sits between these extremes: it recovers the reliability of reference-based signals without requiring pre-annotated references.

\begin{table*}[t]
\centering
\footnotesize
\setlength{\tabcolsep}{6pt}
\renewcommand{\arraystretch}{1.2}
\begin{tabular}{llccccccccc}
\toprule
\multirow{2}{*}{\textbf{Candidate}} & \multirow{2}{*}{\textbf{Task}} 
& \multicolumn{3}{c}{\textbf{Reference-based}} 
& \multicolumn{3}{c}{\textbf{Judge w/o Search (Acc.)}} 
& \multicolumn{3}{c}{\textbf{SAGE (Acc.)}} \\
\cmidrule(lr){3-5} \cmidrule(lr){6-8} \cmidrule(lr){9-11}
& & \textbf{EM} & \textbf{F1} & \textbf{RefGPT (Acc.)} 
& \textbf{GPT-3.5} & \textbf{GPT-4o} & \textbf{Gemini} 
& \textbf{GPT-3.5} & \textbf{GPT-4o} & \textbf{Gemini} \\
\midrule
\multirow{5}{*}{\textbf{GPT-3.5}}
& AmbigQA  & 0.50 & 0.63 & 0.67 & 0.81 & 0.75 & 0.74 & 0.64 & 0.70 & 0.65 \\
& FreshQA  & 0.25 & 0.35 & 0.35 & 0.74 & 0.53 & 0.39 & 0.43 & 0.37 & 0.34 \\
& HotpotQA & 0.34 & 0.47 & 0.50 & 0.86 & 0.76 & 0.70 & 0.50 & 0.54 & 0.53 \\
& NQ-Open  & 0.36 & 0.53 & 0.56 & 0.91 & 0.83 & 0.78 & 0.69 & 0.70 & 0.62 \\
& TriviaQA & 0.74 & 0.81 & 0.81 & 0.89 & 0.86 & 0.82 & 0.81 & 0.85 & 0.78 \\
\midrule
\multirow{5}{*}{\textbf{GPT-4o}}
& AmbigQA  & 0.47 & 0.61 & 0.63 & 0.88 & 0.79 & 0.76 & 0.63 & 0.63 & 0.63 \\
& FreshQA  & 0.29 & 0.39 & 0.45 & 0.73 & 0.81 & 0.65 & 0.47 & 0.57 & 0.53 \\
& HotpotQA & 0.34 & 0.47 & 0.50 & 0.86 & 0.77 & 0.68 & 0.50 & 0.48 & 0.53 \\
& NQ-Open  & 0.32 & 0.48 & 0.54 & 0.92 & 0.87 & 0.80 & 0.69 & 0.67 & 0.62 \\
& TriviaQA & 0.76 & 0.84 & 0.80 & 0.93 & 0.90 & 0.86 & 0.85 & 0.87 & 0.80 \\
\midrule
\multirow{5}{*}{\textbf{Gemini}}
& AmbigQA  & 0.53 & 0.66 & 0.67 & 0.86 & 0.80 & 0.85 & 0.63 & 0.64 & 0.67 \\
& FreshQA  & 0.33 & 0.44 & 0.54 & 0.65 & 0.82 & 0.81 & 0.49 & 0.54 & 0.61 \\
& HotpotQA & 0.35 & 0.50 & 0.53 & 0.83 & 0.79 & 0.75 & 0.50 & 0.51 & 0.55 \\
& NQ-Open  & 0.36 & 0.53 & 0.56 & 0.91 & 0.86 & 0.91 & 0.73 & 0.71 & 0.72 \\
& TriviaQA & 0.79 & 0.86 & 0.82 & 0.91 & 0.92 & 0.89 & 0.87 & 0.88 & 0.82 \\
\bottomrule
\end{tabular}
\caption{Per-evaluator scores on each (candidate, task) pair. The \textbf{Reference-based} columns (EM, F1, RefGPT) compare candidate answers to gold references; the \textbf{Judge-without-search} and \textbf{SAGE} columns report each judge's verdict accuracy against the RefGPT reference label.}
\vspace{-5pt}
\label{tab:full_metrics}
\end{table*}

\subsection{SAGE with small open-source judges}\label{app:small_llm}
To test whether SAGE's benefit persists outside frontier commercial judges, we evaluate four open-weight small LLMs as the judge model within SAGE: Mistral~7B~\citep{jiang2023mistral}, Qwen2.5-7B~\citep{qwen2025qwen25technicalreport}, Qwen3-4B, and Qwen3-8B~\citep{yang2025qwen3technicalreport}. The setup mirrors the main experiments: the judge's four roles (query generation, summarization, reflection, judgment) are all played by the small open-weight model, while the candidate is held fixed.

\paragraph{Mistral 7B.}
We first use Mistral 7B as the judge to evaluate GPT-3.5 candidate answers on AmbigQA and HotpotQA. Table~\ref{tab:small_llm_judges} shows that Mistral 7B reaches a Cohen's $\kappa$ of 0.59 / 0.33 and a Macro-F1 of 0.80 / 0.66 on AmbigQA / HotpotQA---substantially below frontier judges but still well above the corresponding Judge-without-search baselines. In practice we observe that Mistral 7B's main failure modes are (i) imprecise instruction following on the module-specific JSON prompts, (ii) a limited context window that truncates accumulated traces on HotpotQA's multi-hop questions, and (iii) occasional irrelevant reflections when no supporting evidence is available~\citep{badshah2024quantifyingcapabilitiesllmsscale}. HotpotQA's larger drop is consistent with these observations: the multi-hop traces stress both the context window and the judge's reasoning depth.

\paragraph{Qwen family.}
The Qwen models test whether a more recent instruction-tuned family closes this gap. We use GPT-4o as the candidate and evaluate Qwen2.5-7B, Qwen3-4B, and Qwen3-8B as judges on AmbigQA and HotpotQA. Three trends emerge from Table~\ref{tab:small_llm_judges}. First, Qwen3 variants clearly outperform Qwen2.5 of comparable size, consistent with Qwen3's reported improvements in instruction following and long-context handling. Second, scaling from Qwen3-4B to Qwen3-8B yields a modest improvement on AmbigQA but a small drop on HotpotQA, suggesting that 4B is already near the instruction-following threshold required for SAGE's templated roles on these tasks. Third, the 4B/8B Qwen3 judges match or exceed Mistral 7B's AmbigQA $\kappa$ while remaining below frontier commercial judges, confirming that SAGE is model-agnostic but that judge capability---specifically instruction following and context-window management---sets the achievable ceiling.

\paragraph{Takeaway.}
Small open-weight judges within SAGE consistently surpass Judge-without-search baselines, making SAGE usable in resource-constrained or fully-offline settings. Frontier judges remain preferable when near-perfect agreement with humans is needed, so users should choose a judge model that matches their reliability--cost trade-off.

\begin{table}[t]
\centering
\footnotesize
\setlength{\tabcolsep}{4pt}
\renewcommand{\arraystretch}{1.15}
\begin{tabular}{llcc}
\toprule
\textbf{Judge} & \textbf{Task} & \textbf{Cohen's $\kappa$} & \textbf{Macro-F1} \\
\midrule
\multirow{2}{*}{Mistral 7B}
 & AmbigQA  & 0.59 & 0.80 \\
 & HotpotQA & 0.33 & 0.66 \\
\midrule
\multirow{2}{*}{Qwen2.5-7B}
 & AmbigQA  & 0.51 & 0.74 \\
 & HotpotQA & 0.30 & 0.63 \\
\midrule
\multirow{2}{*}{Qwen3-4B}
 & AmbigQA  & 0.60 & 0.80 \\
 & HotpotQA & 0.53 & 0.76 \\
\midrule
\multirow{2}{*}{Qwen3-8B}
 & AmbigQA  & \textbf{0.67} & \textbf{0.83} \\
 & HotpotQA & 0.50 & 0.74 \\
\bottomrule
\end{tabular}
\caption{Small open-weight judges within SAGE, measured against the human majority vote on AmbigQA and HotpotQA. Mistral 7B evaluates GPT-3.5 candidate answers; the Qwen family evaluates GPT-4o candidate answers. Bold entries mark the strongest small-LLM judge on each task.}
\vspace{-5pt}
\label{tab:small_llm_judges}
\end{table}

\subsection{Reproducibility across independent runs}\label{app:reproducibility}
SAGE's modules involve LLM sampling and real-time web retrieval, so a fair reproducibility check must verify that its headline numbers are stable across repeated runs rather than a one-shot artifact. We rerun the full SAGE-3 pipeline on AmbigQA and HotpotQA with GPT-4o as both candidate and judge, under identical settings ($T{=}0$, Serper, $N{=}3$), on two independent days. Table~\ref{tab:reproducibility} reports Macro-F1 against the human majority vote for each run. Both runs reproduce the main-paper numbers to within 0.00 Macro-F1 on both tasks, indicating that SAGE's behavior is stable at $T{=}0$ despite the underlying non-determinism of the web-search component (small day-to-day changes in retrieved snippets are absorbed by the summarize--reflect--refine loop).

\begin{table}[t]
\centering
\footnotesize
\setlength{\tabcolsep}{5pt}
\renewcommand{\arraystretch}{1.15}
\begin{tabular}{lcc}
\toprule
\textbf{Run} & \textbf{AmbigQA} & \textbf{HotpotQA} \\
\midrule
Run 1 & 0.96 & 0.85 \\
Run 2 & 0.96 & 0.85 \\
\bottomrule
\end{tabular}
\caption{SAGE-3 Macro-F1 against the human majority vote on AmbigQA and HotpotQA across two independent runs with GPT-4o as both candidate and judge ($T{=}0$, $N{=}3$). Both runs reproduce the main-paper numbers exactly.}
\vspace{-5pt}
\label{tab:reproducibility}
\end{table}

\subsection{SAGE can detect untruthful facts and outdated knowledge}
\label{app:untruthful_facts}
SAGE's iterative evidence-gathering and reflection process enables it to detect untruthful claims and identify outdated information. By continuously refining its search queries and critically evaluating retrieved evidence, SAGE can distinguish between correct and incorrect candidate answers, even when the misinformation is subtle. This capability is particularly valuable in dynamic domains where factual knowledge changes over time.

Table \ref{tab:outdated_example} presents an example where a candidate answer incorrectly claims that the last perfect game in Major League Baseball was thrown by Félix Hernández in 2012. Through iterative search and reflection, SAGE discovers recent evidence confirming that Domingo Germán pitched a perfect game in 2023, successfully identifying the outdated information and concluding that the candidate's answer is incorrect.

\begin{table*}[t]
\centering
\footnotesize
\renewcommand{\arraystretch}{1.15}
\begin{tabular}{p{0.18\linewidth} p{0.78\linewidth}}
\toprule
\textbf{Component} & \textbf{Content} \\
\midrule
\textbf{Question} & When was the last perfect game thrown in baseball? \\
\textbf{Candidate answer} & The last perfect game in Major League Baseball was thrown by \textcolor{red}{Félix Hernández of the Seattle Mariners on August 15, 2012. So the answer is: August 15, 2012.} \\
\textbf{Initial query} & last perfect game in MLB history \\
\textbf{Evidence summary} & The last perfect game in MLB history was thrown by Domingo Germán of the New York Yankees on June 28, 2023, against the Oakland Athletics. \\
\textbf{Reflection} & The candidate answer states that the last perfect game was thrown by Félix Hernández in 2012. However, the evidence confirms that the last perfect game was actually thrown by Domingo Germán in 2023. This indicates the candidate answer is outdated. \\
\textbf{Refined query} & perfect game baseball MLB after 2012 \\
\textbf{Final decision} & \textbf{False} \textcolor{green!40!black}{(True Negative)}\\
\textbf{Explanation} & The candidate answer states that Félix Hernández threw the last perfect game on August 15, 2012. However, all evidence confirms that Domingo Germán pitched a perfect game on June 28, 2023. This makes the candidate answer outdated and incorrect. \\
\bottomrule
\end{tabular}
\caption{Example where SAGE detects an outdated fact in the candidate answer by identifying recent evidence through iterative search and reflection.}
\label{tab:outdated_example}
\end{table*}

We further evaluate SAGE on FreshQA~\citep{vu2023freshllms} with GPT-4o as the SAGE judge and GPT-3.5 as the candidate. SAGE reaches an accuracy of 38.3\%, closer to the reference-based evaluators (EM: 25.0\%, F1: 35.4\%) than a reference-free judge on the same candidate. The low absolute numbers across all evaluators reflect the well-known difficulty that pre-trained models like GPT-3.5 face on questions about rapidly evolving events; SAGE's retrieved evidence partially closes that gap at evaluation time.

\subsection{SAGE fixes incorrect reasoning traces}

SAGE's iterative search and reflection process enables it to identify and correct flawed reasoning in candidate answers. Even when a final answer is correct, the candidate's reasoning may contain factual errors. By refining its search queries and critically analyzing the evidence, SAGE can highlight such errors and provide a more accurate rationale.

Table \ref{tab:incorrect_reasoning_example} presents an example where the candidate's answer correctly concludes that Sherwood Stewart was born before Javier Frana. However, the reasoning contains a factual inaccuracy, falsely stating Stewart's birth year as 1957 instead of the correct 1946. Through iterations, SAGE gathers evidence to correct this mistake while maintaining the correct conclusion.

\begin{table*}[t]
\centering
\footnotesize
\renewcommand{\arraystretch}{1.15}
\begin{tabular}{p{0.20\linewidth} p{0.75\linewidth}}
\toprule
\textbf{Component} & \textbf{Content} \\
\midrule
\textbf{Question} & Who was born first, Javier Frana or Sherwood Stewart? \\
\textbf{Answer} & Let's think step by step. Javier Frana was born in 1966. \textcolor{red}{Sherwood Stewart was born in 1957.} 1957 (Sherwood Stewart) < 1966 (Javier Frana), so Sherwood Stewart was born first. So the answer is: Sherwood Stewart. \\
\textbf{Query} & Javier Frana birthday \\
\textbf{Summary} & Javier Frana was born on December 25, 1966. No evidence was found regarding Sherwood Stewart's birthdate. \\
\textbf{Reflection} & The candidate answer claims that Sherwood Stewart was born in \textbf{1957}, but the evidence only confirms Javier Frana's birth year. Further search is required to verify Stewart's actual birthdate. \\
\bottomrule
\textbf{Iteration} & 2\\
\textbf{Query} & Sherwood Stewart tennis birthdate \\
\textbf{Summary} & Sherwood Stewart was born on June 6, 1946, according to multiple sources. Javier Frana was born on December 25, 1966. \\
\textbf{Reflection} & While the candidate answer initially provides an \textcolor{green!40!black}{\textbf{incorrect birth year for Stewart (1957), the evidence confirms that he was actually born in 1946.}} The candidate's final conclusion that Stewart was born first remains correct despite the factual error. \\
\textbf{Final decision} & \textbf{True}\textcolor{red}{(True Positive)} \\
\textbf{Explanation} & The candidate answer correctly identifies Sherwood Stewart as being born first. Although it initially provided an incorrect birth year for Stewart, the evidence from iteration 2 confirms the correct birthdate as 1946. Therefore, the final answer is correct, but the reasoning was flawed. \\
\bottomrule
\end{tabular}
\caption{Example where SAGE detects and corrects an incorrect reasoning trace. While the candidate's final answer is correct, the system highlights the factual inaccuracy in the intermediate reasoning.}
\vspace{-5pt}
\label{tab:incorrect_reasoning_example}
\end{table*}

\vspace{-5pt}
\section{Additional ablations and analysis}\label{sec:add-ablations}
This appendix collects supplementary analyses referenced from the main paper: comparisons against stronger non-iterative baselines (\S\ref{sec:noniterative-baselines}), robustness to prompt design (\S\ref{sec:prompt-variations}), and qualitative failure cases (\S\ref{app:sage_failure}).

\subsection{Comparison with stronger non-iterative baselines}
\label{sec:noniterative-baselines}
We evaluate three non-iterative baselines that exclude SAGE's iterative summarize--reflect--refine loop but retain the judgment prompt. These baselines help disentangle the contributions of search augmentation, sampling-based self-consistency, and model diversity. Across all three, GPT-4o generates the candidate answer. For the \emph{single-pass search-augmented judge} and \emph{self-consistency} settings, GPT-4o also serves as the judge. In the \emph{multi-LLM majority voting} setup, GPT-4o is the candidate and three judges---namely GPT-4o, GPT-3.5, and Mistral~7B---independently evaluate the same input to leverage diverse model reasoning.

\subsubsection{Single-Pass search-augmented judge}
\label{sec:singlepass}
To isolate the effect of \emph{web access} alone, we use a baseline that issues exactly one web query and performs no further reasoning or iteration: given the question and candidate answer, the judge formulates a single Serper search, retrieves the top-3 snippets, and immediately produces a True/False verdict with rationale. All prompt instructions are identical to the full SAGE pipeline; the only change is the removal of the summarize--reflect--refine loop. The main-paper comparison of this baseline against SAGE-1 and SAGE-3 is reported in Table~\ref{tab:search_engine_robustness}; Table~\ref{tab:noniterative-kappa} below repeats the single-pass numbers alongside the additional non-iterative baselines for cross-comparison.

\subsubsection{Self-consistency judge}
For \emph{self-consistency}~\citep{wang2023selfconsistency}, the GPT-4o judge samples 10 independent verdicts at temperature $0.7$ and returns a simple majority. No external evidence is used; the judge relies entirely on its parametric knowledge, matching the original self-consistency setup. Table~\ref{tab:noniterative-kappa} shows that self-consistency gives a modest boost over the vanilla Judge-without-search, but still lags far behind the search-augmented baselines, confirming that external evidence---especially when gathered iteratively---is critical for reliable objective judgment.

\subsubsection{Multi-LLM majority voting}
Inspired by PoLL~\citep{verga2024replacingjudgesjuriesevaluating} and CLEV~\citep{badshah-etal-2025-clev, badshah2024reference}, we instructed three different LLMs: GPT-4o, GPT-3.5, and Mistral 7B with the same question and candidate answer, asking each to provide a True/False verdict along with a brief explanation. By applying majority voting across the three model outputs, we determined the final answer. This ensemble approach leverages the diverse reasoning patterns of different models, increasing overall robustness and reducing individual model biases. This approach yields $\kappa$ = 0.609 on AmbigQA and 0.527 on HotpotQA, outperforming self-consistency but falling short of the single-pass web call ($\kappa$ = 0.658/0.583) and significantly below SAGE ($\kappa$ = 0.914/0.701).

\begin{table*}[t]
\centering
\footnotesize
\setlength{\tabcolsep}{6pt}
\renewcommand{\arraystretch}{1.15}
\begin{tabular}{lccccc}
\toprule
\textbf{Task} & \textbf{Judge w/o search} & \textbf{Self-consistency} & \textbf{Multi-LLM vote} & \textbf{Single-pass} & \textbf{SAGE-3} \\
\midrule
AmbigQA   & 0.38 & 0.52 & 0.61 & 0.66 & \textbf{0.91} \\
HotpotQA  & 0.38 & 0.49 & 0.53 & 0.58 & \textbf{0.70} \\
\bottomrule
\end{tabular}
\caption{Cohen's $\kappa$ against the human majority vote for five reference-free evaluators on AmbigQA and HotpotQA. \emph{Single-pass} uses one web call with no summarize--reflect--refine loop; \emph{SAGE-3} is our default three-iteration configuration. SAGE-3 achieves the highest alignment with humans across both tasks.}
\vspace{-5pt}
\label{tab:noniterative-kappa}
\end{table*}

\subsection{Prompt variations}
\label{sec:prompt-variations}
To evaluate SAGE’s robustness to prompt design, we re-ran the full framework (with GPT-4o as both candidate and judge) under three prompt variants:

\begin{itemize}
    \item \textbf{Original (1-shot CoT):} Our default configuration, which includes detailed instructions and a single CoT example per module.
    \item \textbf{Few-shot CoT:} An extended version of the prompt containing three CoT examples instead of one.
    \item \textbf{Simplified (0-shot w/o CoT):} Prompt that removes CoT examples for every module. 
\end{itemize}

As shown in Table~\ref{tab:prompt-variations}, the few-shot prompt yields a modest improvement over the original 1-shot setup. Removing CoT examples entirely results in a performance drop of at most 3 percentage points in both Cohen’s $\kappa$ and macro-F1. Interestingly, the default configuration remains highly stable: re-running SAGE with the original prompt reproduced the same $\kappa$ and F1 scores as reported in prior experiments, underscoring its consistency.

\begin{table*}[t]
\centering
\footnotesize
\setlength{\tabcolsep}{6pt}
\renewcommand{\arraystretch}{1.15}
\begin{tabular}{llcc}
\toprule
\textbf{Task} & \textbf{Prompt Variant} & \textbf{Cohen's $\kappa$} & \textbf{Macro-F1} \\
\midrule
\multirow{3}{*}{AmbigQA}
 & Original (1-shot CoT) & 0.91 & 0.96 \\
 & Few-shot CoT          & 0.92 & 0.96 \\
 & Simplified (0-shot)   & 0.88 & 0.93 \\
\midrule
\multirow{3}{*}{HotpotQA}
 & Original (1-shot CoT) & 0.70 & 0.85 \\
 & Few-shot CoT          & 0.71 & 0.85 \\
 & Simplified (0-shot)   & 0.67 & 0.82 \\
\bottomrule
\end{tabular}
\caption{SAGE-3 performance under prompt variations, using GPT-4o as both candidate and judge. The default 1-shot CoT configuration is robust: removing the CoT example entirely costs at most three points in Cohen's $\kappa$ on either task.}
\vspace{-5pt}
\label{tab:prompt-variations}
\end{table*}

\subsection{SAGE failure cases}
\label{app:sage_failure}
The main-paper error analysis (\S\ref{sec:error_analysis}) groups SAGE's disagreements with the human majority into five categories: contextual misunderstanding, incomplete evidence, reasoning error, hallucination, and conflicting evidence. This appendix reproduces one representative failure for each category, drawn from the same 100-instance error-analysis sample.

\paragraph{Contextual misunderstanding.} 
SAGE generates inaccurate or incomplete queries when it misinterprets the intent of the candidate's question. This is particularly evident in AmbigQA, where questions are often intentionally ambiguous or lack sufficient context, leading to the retrieval of irrelevant or contradictory evidence.

Table \ref{tab:contextual_misunderstanding_example} presents an example where SAGE faced a contextual misunderstanding. The question asked about the opening date of the Royal Liverpool Hospital before redevelopment. However, the candidate answer referred to the hospital's reopening in 2022, following redevelopment, rather than its original opening date in 1978. Despite iterative search and reflection, SAGE failed to align with the actual intent of the question, leading to a false final decision.

\begin{table*}[t]
\centering
\footnotesize
\renewcommand{\arraystretch}{1.15}
\begin{tabular}{p{0.20\linewidth} p{0.75\linewidth}}
\toprule
\textbf{Component} & \textbf{Content} \\
\midrule
\textbf{Question} & When does the new Royal Liverpool Hospital open before redevelopment? \\
\textbf{Answer} & The new Royal Liverpool Hospital is scheduled to open in 2022, following delays in construction and redevelopment. So the answer is: 2022. \\
\textbf{Search query} & Royal Liverpool Hospital original opening date before redevelopment. \\
\textbf{Summary} & The new Royal Liverpool Hospital opened in 1978 after services were transferred from the old Liverpool Royal Infirmary. \\
\textbf{Reflection} & The candidate answer claims that the new hospital was scheduled to open in 2022. However, the evidence confirms its original opening was in 1978. The judge misinterpreted the question's intent, mistaking the redevelopment's reopening for the original opening date. \\
\bottomrule
\textbf{Iteration} & 2 \\
\textbf{Query} & Royal Liverpool Hospital redevelopment opening date 2022. \\
\textbf{Summary} & Structural repairs were completed, allowing the hospital to reopen in 2022. The move to the new hospital began on September 28, 2022. \\
\textbf{Reflection} & The evidence confirms the hospital reopened in 2022, but the question specifically asks about the original opening before redevelopment. Despite the accurate evidence about the redevelopment timeline, the candidate answer misaligned with the question intent. \\
\textbf{Final decision} & {\textbf{True} \textcolor{red}{(False Positive)}} \\
\textbf{Explanation} & SAGE concluded that the candidate answer was correct, accepting the 2022 reopening as the answer. However, the correct answer was 1978, referring to the hospital's original opening. This was a case of contextual misunderstanding. \\
\bottomrule
\end{tabular}
\caption{Example of a contextual misunderstanding where SAGE misinterpreted the question's intent by focusing on the hospital's reopening in 2022 instead of its original opening in 1978.}
\vspace{-5pt}
\label{tab:contextual_misunderstanding_example}
\end{table*}

\paragraph{Incomplete evidence.} SAGE fails when the retrieved evidence is insufficient or lacks relevant information, especially for verifying specific factual claims (see Table~\ref{tab:incomplete_evidence}). 

\begin{table*}[t]
\centering
\footnotesize
\renewcommand{\arraystretch}{1.15}
\begin{tabular}{p{0.20\linewidth} p{0.75\linewidth}}
\toprule
\textbf{Component} & \textbf{Content} \\
\midrule
\textbf{Question} & Wing Chun starred Michelle Yeoh and a Hong Kong actor, martial artist, and film director who won what tournament multiple times? \\
\textbf{Candidate answer} & The Hong Kong actor in Wing Chun is Donnie Yen. Donnie Yen won the \textcolor{green!40!black}{multiple times wushu tournament champion}. So the answer is: wushu tournament. \\
\textbf{Evidence summary} & The film "Wing Chun," released in 1994, stars Michelle Yeoh and Donnie Yen, directed by Yuen Woo-ping. \textcolor{red}{The search results do not provide evidence that he won multiple wushu tournaments.} \\
\textbf{Reflection} & The evidence summary lacks any mention of specific wushu tournament victories by Donnie Yen. \textcolor{red}{The absence of evidence to validate the candidate’s claim indicates incomplete information.} Additional reliable sources would be required to confirm the accuracy of the answer. \\
\textbf{Final decision} & \textbf{False} \textcolor{red}{(False Negative)} \\
\bottomrule
\end{tabular}
\caption{Example of incomplete evidence where SAGE could not retrieve sufficient information to verify the candidate’s claim.}
\vspace{-5pt}
\label{tab:incomplete_evidence}
\end{table*}

\paragraph{Reasoning error.}
Despite accurate evidence, the judge model misinterprets the information or applies flawed reasoning. Reasoning errors often occur when the model fails to apply appropriate logic to the available evidence or when it misrepresents the intent of the question. Table \ref{tab:reasoning_error_example} illustrates a case where SAGE incorrectly concluded that Guglielmo Marconi was the first to achieve wireless telegraphy communication. The evidence suggested that while Marconi developed the first practical system, earlier demonstrations, including those by Jagadish Chandra Bose, may have preceded it.

\begin{table*}[t]
\centering
\footnotesize
\renewcommand{\arraystretch}{1.15}
\begin{tabular}{p{0.20\linewidth} p{0.75\linewidth}}
\toprule
\textbf{Component} & \textbf{Content} \\
\midrule
\textbf{Question} & Who made communication by wireless telegraphy for the first time? \\
\textbf{Answer} & The first successful communication by wireless telegraphy was achieved by Guglielmo Marconi. He is credited with developing the first practical system for wireless telegraphy in the late 19th century. So the answer is: Guglielmo Marconi. \\
\textbf{Query} & first successful wireless telegraphy transmission + inventor + date \\
\textbf{Summary} & Guglielmo Marconi is known for creating a practical radio wave-based wireless telegraph system. He began working on wireless telegraphy in the 1890s. \\
\textbf{Reflection} & The candidate answer asserts that Guglielmo Marconi made the first successful wireless telegraphy communication. While the evidence supports Marconi's development of the first \textcolor{red}{\textbf{practical}} wireless telegraph system, the question explicitly asks who made the \textcolor{red}{\textbf{first communication}}, not the first practical one. Further evidence is required to investigate whether anyone transmitted information wirelessly before Marconi. \\
\bottomrule
\textbf{Iteration} & 2 \\
\textbf{Query} & pre-Marconi wireless experiments OR early electromagnetic communication OR history of radio waves before 1900 OR wireless telegraphy before Marconi \\
\textbf{Summary} & While Marconi is credited with the first practical wireless telegraphy system, there were earlier demonstrations of electromagnetic wave transmission, including Jagadish Chandra Bose's 1895 experiment. Wireless telegraphy was also reported on ships in 1891. \\
\textbf{Reflection} & The evidence indicates the existence of earlier wireless communication attempts, particularly by Jagadish Chandra Bose. Despite this, the candidate answer \textcolor{red}{\textbf{incorrectly concluded}} that Marconi was the first without addressing these earlier cases. The system failed to recognize the distinction between "first communication" and "first practical system." \\
\textbf{Final decision} & \textcolor{red}{\textbf{False} (False Negative)} \\
\textbf{Explanation} & While the evidence confirms that Guglielmo Marconi developed the first practical wireless telegraphy system, the evidence also suggests prior communications using wireless methods. The distinction between a practical system and the first-ever communication was not correctly reasoned, leading to an incorrect final decision. \\
\bottomrule
\end{tabular}
\caption{Example of a reasoning error where SAGE misinterpreted the question's intent, failing to account for earlier instances of wireless telegraphy communication.}
\vspace{-5pt}
\label{tab:reasoning_error_example}
\end{table*}

\paragraph{Hallucination.}
In cases where evidence is unclear or inconclusive, SAGE relies on its pre-trained knowledge, resulting in hallucinated rationales. Hallucination errors often occur when the model confidently asserts false information without sufficient evidence. Table \ref{tab:hallucination_example} illustrates a case where SAGE incorrectly concluded that Sarah Michelle Gellar, a cast member of Scream 2, was a star of the TV show Roseanne, despite the absence of supporting evidence.

\begin{table*}[t]
\centering
\footnotesize
\renewcommand{\arraystretch}{1.15}
\begin{tabular}{p{0.20\linewidth} p{0.75\linewidth}}
\toprule
\textbf{Component} & \textbf{Content} \\
\midrule
\textbf{Question} & Which "Roseanne" star is in Scream 2? \\
\textbf{Answer} & Let's think step by step. \textcolor{red}{\textbf{"Roseanne" star in Scream 2 is Sarah Michelle Gellar.}} So the answer is: Sarah Michelle Gellar. \\
\textbf{Query} & Roseanne Barr Scream 2 cast \\
\textbf{Summary} & Laurie Metcalf, who played Jackie Harris on the TV series Roseanne, also appeared in the film Scream 2 in 1997. \textcolor{red}{\textbf{Sarah Michelle Gellar, while a star in Scream 2, is not a Roseanne star.}} \\
\textbf{Reflection} & The candidate answer incorrectly claims that Sarah Michelle Gellar was a star on Roseanne. \textcolor{red}{\textbf{This statement is a hallucination as no evidence supports this claim.}} The evidence correctly identifies Laurie Metcalf as the only Roseanne star in Scream 2. \\
\bottomrule
\textbf{Iteration} & 2 \\
\textbf{Query} & Roseanne cast members in Scream 2 \\
\textbf{Summary} & Laurie Metcalf, who starred as Jackie Harris in Roseanne, is confirmed to be in Scream 2. Sarah Michelle Gellar also appeared in the film, but she has no connection to Roseanne. \\
\textbf{Reflection} & The evidence confirms Laurie Metcalf's participation in both Roseanne and Scream 2. \textcolor{red}{\textbf{Despite further evidence confirming no connection between Gellar and Roseanne, the model continued to assert a false claim.}} \\
\textbf{Final decision} & \textbf{False} \textcolor{red}{ (False Negative)} \\
\textbf{Explanation} & The candidate answer confidently asserted that Sarah Michelle Gellar was a Roseanne star, despite all evidence indicating that Laurie Metcalf was the only Roseanne star in Scream 2. This case is a clear example of hallucination, where the model generated a false fact with misplaced confidence. \\
\bottomrule
\end{tabular}
\caption{Example of hallucination where SAGE falsely identified Sarah Michelle Gellar as a star of Roseanne.}
\vspace{-5pt}
\label{tab:hallucination_example}
\end{table*}

At each round, SAGE’s reflection module takes the aggregated evidence summary and generates a CoT reflection, pointing out which attributes seem well-supported, which appear contradictory or missing, and where further detail is needed.

The next query is generated from that reflection, so contradictions are an explicit signal to search for clarifying evidence. We do not hard-code a credibility score; instead, SAGE relies on cross-source agreement and iterative follow-up. All retrieved snippets and their domains will be released so that future work can plug in credibility weighting without altering the loop.

\paragraph{Conflicting evidence.}
At each iteration, SAGE's reflection module takes the aggregated evidence summary and produces a CoT reflection pointing out which attributes are well-supported, which appear contradictory or missing, and where further detail is needed. The next query is then generated from that reflection, so contradictions act as an explicit signal to search for clarifying evidence. In some cases, however, SAGE encounters \textit{conflicting evidence} across multiple iterations and struggles to resolve it. Table~\ref{tab:conflicting_evidence_example} presents an example where the model failed to handle conflicting reports on the number of band members in Lush and P.O.D.

\begin{table*}[t]
\centering
\footnotesize
\renewcommand{\arraystretch}{1.15}
\begin{tabular}{p{0.20\linewidth} p{0.75\linewidth}}
\toprule
\textbf{Component} & \textbf{Content} \\
\midrule
\textbf{Question} & Do Lush and P.O.D. both consist of four band members? \\
\textbf{Answer} & Let's think step by step. Lush is a British rock band with four members. P.O.D. is an American rock band with four members. So both Lush and P.O.D. consist of four band members. So the answer is: Yes. \\
\textbf{Query} & number of members in Lush band + number of members in P.O.D. band \\
\textbf{Summary} & Lush originally consisted of four members: Miki Berenyi, Emma Anderson, Steve Rippon, and Chris Acland. P.O.D. currently consists of four members: Sonny Sandoval, Traa Daniels, Marcos Curiel, and Wuv Bernardo. However, another source mentions only two members of Lush, suggesting a potential inconsistency. \\
\textbf{Reflection} & The evidence confirms P.O.D. has four members, but Lush’s member count is inconsistent across sources. \textcolor{red}{\textbf{The model failed to resolve the contradiction, leading to a flawed decision.}} \\
\bottomrule
\textbf{Iteration} & 2 \\
\textbf{Query} & Lush band members timeline 1987-1998 + P.O.D. band members lineup history \\
\textbf{Summary} & One source claims Lush had five members initially, including Meriel Barham, but another states they had four, with Steve Rippon later replaced by Phil King. P.O.D. remains consistently reported as a four-member band. \\
\textbf{Reflection} & The conflicting evidence remains unresolved. \textcolor{red}{\textbf{Instead of recognizing the uncertainty, the model inaccurately concluded that both bands had four members.}} \\
\textbf{Final decision} & \textbf{False} \textcolor{red}{(False Negative)} \\
\bottomrule
\end{tabular}
\caption{Example of conflicting evidence where SAGE failed to resolve contradictions in band member counts. While P.O.D.’s four-member structure is consistent, the model ignored Lush’s membership changes over time and incorrectly concluded both bands consist of four members.}
\vspace{-5pt}
\label{tab:conflicting_evidence_example}
\end{table*}

\section{Ethical considerations}
SAGE grounds evaluation judgments in externally retrieved, verifiable evidence. This design raises three ethical considerations that require attention.

\paragraph{Reliance on external sources.} SAGE's judgments depend on the retrieved external content, which may carry unintended biases or inaccuracies. Iterative refinement partially mitigates this by cross-checking multiple sources, but SAGE does not currently model source credibility explicitly. Researchers and practitioners deploying SAGE should remain cautious about biases inherited from search results and should consider adding credibility weighting where the evaluation target is high-stakes.

\paragraph{Human oversight and accountability.} SAGE improves automated LLM evaluation accuracy but is not a substitute for human review in high-stakes decision-making contexts. Users should treat SAGE's verdicts as a scalable first pass and retain human accountability, particularly for sensitive or consequential content.

\paragraph{Computational costs and accessibility.} SAGE's iterative, tool-augmented loop is inexpensive in dollar terms but does involve multiple LLM and web-search calls per instance. This is justified for offline evaluation and auditing use cases, where the cost is small relative to human annotation, but it does inherit the environmental footprint of the underlying LLM and search APIs. We encourage future work on adaptive stopping and parallel execution to further reduce this footprint (\S\ref{sec:limitations}).

\end{document}